\documentclass[lettersize,journal]{IEEEtran}
\usepackage{amsmath,amsfonts}
\usepackage{algorithmic}
\usepackage[ruled]{algorithm2e}
\usepackage{array}
\usepackage[caption=false,font=normalsize,labelfont=sf,textfont=sf]{subfig}
\usepackage{textcomp}
\usepackage{stfloats}
\usepackage{url}
\usepackage{verbatim}
\usepackage{graphicx}
\usepackage{cite}
\usepackage{booktabs}
\usepackage{multirow}
\usepackage{color}
\usepackage{forloop}
\begin{document}
    \title{FedSR: A Semi-Decentralized Federated Learning Algorithm for Non-IIDness in IoT System}
    \author{Jianjun Huang, Lixin Ye, Li Kang}
    \markboth{Journal of \LaTeX\ Class Files,~Vol.~18, No.~9, September~2022}{}
    \maketitle
    \begin{abstract}
    In the Industrial Internet of Things (IoT), a large amount of data will be generated every day. Due to privacy 
and security issues, it is difficult to collect all these data together to train deep learning models, thus the federated learning, a distributed machine 
learning paradigm that protects data privacy, has been widely used in IoT. However, in practical federated learning, the data 
distributions usually have large differences across devices, and the heterogeneity of data will deteriorate the performance of 
the model. Moreover, federated learning in IoT usually has a large number of devices involved in training, and the limited 
communication resource of cloud servers become a bottleneck for training. To address the above issues, in this paper, we combine 
centralized federated learning with decentralized federated learning to design a semi-decentralized cloud-edge-device hierarchical 
federated learning framework, which can mitigate the impact of data heterogeneity, and can be deployed at lage scale in IoT. To 
address the effect of data heterogeneity, we use an incremental subgradient optimization algorithm in each ring cluster to improve 
the generalization ability of the ring cluster models. Our extensive experiments show that our approach can effectively mitigate 
the impact of data heterogeneity and alleviate the communication bottleneck in cloud servers.

\end{abstract}
    \section{Introduction}
    \IEEEPARstart{I}n recent years, deep learning has made tremendous growth in the fields of computer vision, natural language 
    processing, and speech recognition\cite{voulodimos2018deep},\cite{chowdhary2020natural},\cite{nassif2019speech}. One of the 
    keys to the great success of deep learning is the large amount of available data. For deep learning, data is equivalent to 
    the energy power required for the operation of a factory. However, in real-world scenarios, data is usually distributed across 
    mobile devices or organizations. Therefore, collecting this dispersed data requires extensive communication overhead and storage 
    resources. Moreover, due to growing privacy concerns and stricter data regulations like General Data Protection Regulation (GDPR)
    \cite{voigt2017eu}, gathering data on a cloud server for centralized training is becoming progressively more challenging 
    \cite{kairouz2021advances}, \cite{li2020federatedChallenges}. To tackle above issues, Google introduced a novel distributed machine 
    learning paradigm in 2016, known as Federated Learning (FL) \cite{mcmahan2017communication}. Federated learning enables the training 
    of an enhanced global model by exchanging model parameters among devices while preserving data privacy. Federated learning has become 
    a important area in machine learning that has gained significant attention and widespread practical application \cite{bonawitz2019towards}, 
    \cite{hard2018federated}, \cite{kaissis2020secure}.
    \begin{figure}[t]
        \centering
        \includegraphics[width=0.5\textwidth]{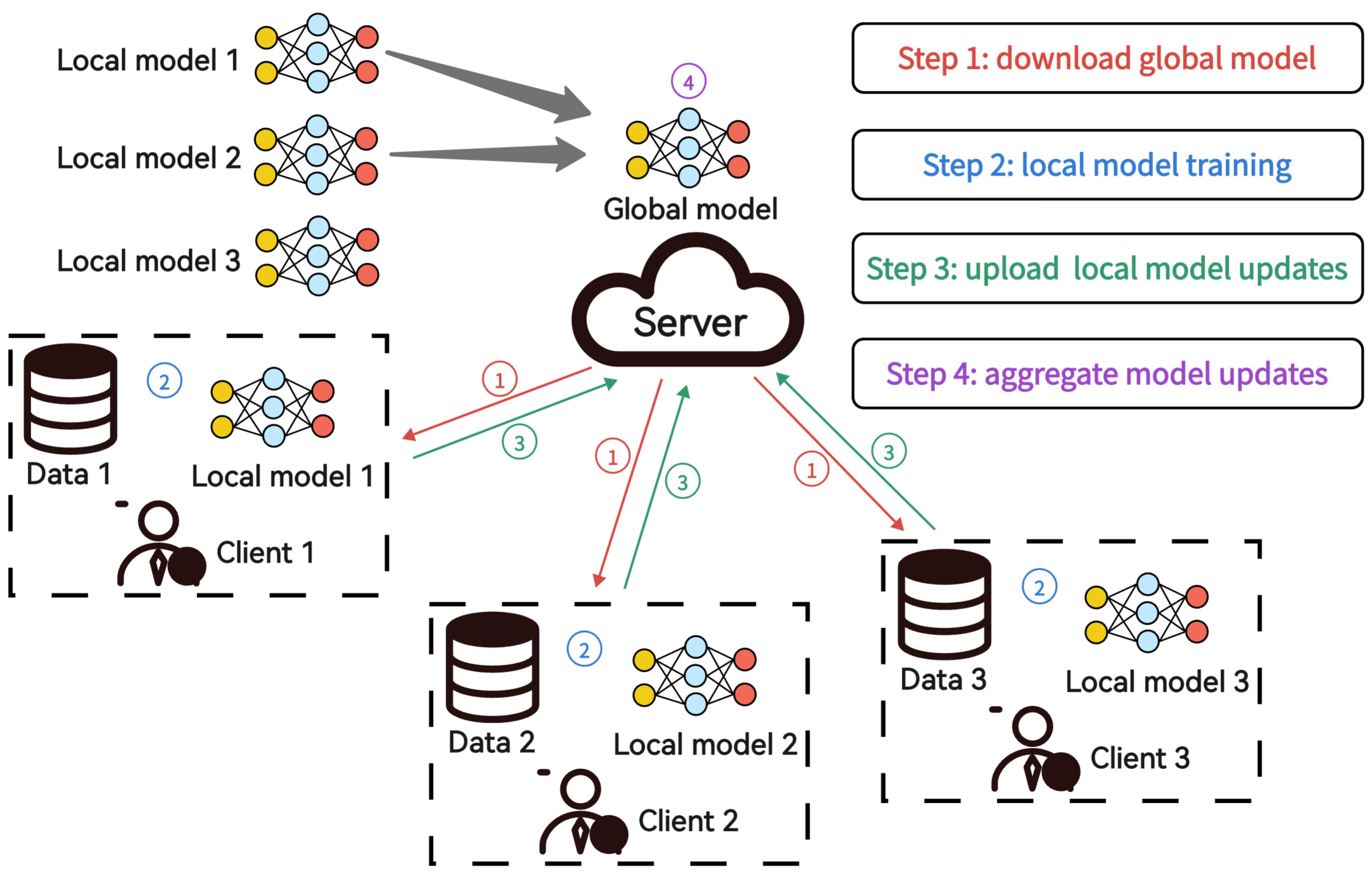}
        \caption{FedAvg framework}
        \label{FedAvg.png}
    \end{figure}

    \par
        Federated learning can be divided into centralized federated learning (CFL) and decentralized federated learning (DFL) according 
    to whether there is a cloud server to assist in training. In CFL, a cloud server is usually available to assist participants to train 
    a global model cooperatively. For example, FedAvg \cite{mcmahan2017communication}, the pioneer of federated learning, the cloud server 
    sends the global model to the participants for local training, and then aggregates the models uploaded by all participants to update 
    the global model and prepare it for the next round of training. While in DFL, the model parameters are shared among the participants 
    for training via peer-to-peer (P2P) manner, without involving the cloud server. The device involved in the training in CFL and DFL only 
    needs to send the local model parameters to the cloud server or other devices, while the original data is kept locally, thus reducing 
    the risk of privacy data leakage. Although the federated learning paradigm mentioned above is promising, it still faces many challenges. 
    For example, the distribution of data in IoT devices is typically heterogeneous, which refers to the presence of non-independent and 
    non-identically distributed (non-iid) data in federated learning. In brief, the data distribution varies significantly across devices. 
    This is because participants may be located in different geographic locations, industries or organizations and have data of various types, 
    formats and characteristics. The study \cite{karimireddy2020scaffold} pointed out that in the non-iid case, the differences between the 
    data distribution of each device and the global data distribution lead to the model trained by each device bias the local optimum. As a 
    result, the global model obtained by the server aggregating the local models of the devices in CFL has deviations from the true optimal 
    solution \cite{karimireddy2020scaffold}. To solve the problem of data heterogeneity, some CFL algorithms have been proposed \cite{li2020federated}, 
    \cite{li2021model}, \cite{karimireddy2020scaffold}. Li et al. \cite{li2020federated} proposed FedProx to mitigate device drift by introducing 
    proximal terms to reduce the difference between the local model and global model. MOON \cite{li2021model} utilizes the contrastive loss 
    function to enhance the consistency between the representations learned by the local model and the global model. SCAFFOLD \cite{karimireddy2020scaffold} 
    introducs a control gradient variate to correct the drift in the local updates. In addition, some studies \cite{jeong2018communication}, 
    \cite{hao2021towards}, \cite{goetz2020federated} mitigate the impact of data heterogeneity by sharing a public dataset or synthesized data 
    to balance the device's data distribution. However, all the above approaches assume that the server has ideal communication resources. The 
    limited communication of the cloud server will be a bottleneck for CFL with large number of devices involved. For DFL, previous research 
    mainly focuses on communication optimization, and there are relatively few studies on solving non-iid problems. Although \cite{li2021decentralized}, 
    \cite{wang2022efficient}, \cite{onoszko2021decentralized} proposed some DFL methods for solving the non-iid problem, there is still a gap in 
    their performance compared to CFL methods.
    \par
        To address data heterogeneity and communication bottlenecks in CFL, we proposed an algorithm called FedSR, a semi-decentralized cloud-edge-device 
    hierarchical federated learning framework with a star-ring topology for IoT applications. In FedSR, firstly, each device will select a nearby edge 
    server to participate in the training. Then, the edge server will randomly connect the participating devices into a ring cluster and an incremental 
    subgradient optimization algorithm \cite{bertsekas2011incremental} is employed to address data heterogeneity. We experimentally demonstrate the 
    effectiveness of the ring incremental algorithm in federated learning with a small number of participating devices. Finally, the cloud server uses 
    the FedAvg algorithm to aggregate the resulting models from each edge server and update the global model. In this way, the cloud server only needs 
    to aggregate the resulting models from each edge server without aggregating the models of all devices in each round, thus alleviating the bottleneck 
    of communication at the cloud server. In addition, in each ring cluster, we adopt the DFL training method, which fully utilizes the communication 
    resources between devices without additional infrastructure costs. In summary, the main contributions of this paper are as follows:
    \begin{enumerate}
        \item{
            We experimentally validate the effectiveness of the incremental subgradient optimization algorithm on federated learning with 
        a small number of devices involved, and it can better solve the problem of data heterogeneity compared to the vanilla federated 
        learning algorithm.
        }
        \item{
            We propose the FedSR algorithm by combining CFL with DFL and an incremental optimization algorithm is employed to tackle data 
        heterogeneity by improving the generalization ability of cluster models in each ring cluster. In FedSR, the cloud server only needs to 
        aggregate the models form edge server instead of the models of all devices, thus reducing the communication pressure on the server. 
        }
        \item{
            We provide a rigorous theoretical analysis of the method proposed in this paper. Our analysis shows that FedSR has provable convergence 
        guarantees and is close to the optimal solution when the learning rate satisfies certain conditions.
        }
        \item{
            We conduct experiments on the FedSR algorithm proposed in this paper on MNIST, FashionMNIST, CIFAR-10, CIFAR-100 datasets. The 
        experimental results show that our proposed algorithm achieves the best model performance than popular federated learning algorithms both 
        in iid and non-iid cases. Additionally, we conducted experiments and analyses on each hyperparameter of the FedSR algorithm, thereby 
        demonstrating the effectiveness of our proposed FedSR algorithm in this paper.
        }
    \end{enumerate}
    \par
        The remainder of this paper is organized as follows. The related work is formulated in Section II. The considered problem, our motivation, 
    the proposed FedSR algorithm and the convergence of FedSR are presented in Section III. In Section IV, we present experimental results to demonstrate the excellent performance 
    of the proposed algorithm compared to the baseline method, and these results are discussed.

    \section{Related Work}
\subsection{Federated Learning on Ring Topology}
    In DFL, several studies have proposed federated learning based on fully connected topology \cite{samarakoon2019distributed}, 
\cite{li2020blockchain}, \cite{zhang2021d2d}. Although this fully-connected DFL avoids the poor scalability and vulnerability to 
single-point attacks of star architectures, it introduces a large amount of redundant communication consumption and makes it 
difficult to ensure model convergence due to the lack of model aggregation \cite{han2023ringffl}. In order to reduce communication 
redundancy and accelerate model convergence, the researchers proposed a ring topology DFL framework \cite{han2023ringffl}, 
\cite{xu2022ring}, \cite{wang2022efficient}, \cite{lee2020tornadoaggregate}. Xu et al. combines the modified Ring All-reduce with 
the modified Ant Colony Optimization algorithm to design a ring topology DFL to mitigate the uplink transmission time of FL in 
wireless networks \cite{xu2022ring}. \cite{wang2022efficient} proposed RDFL (Ring decentralized federated learning) to increase 
the bandwidth utilization as well as the training stability. In RDFL, the clients in the ring topology will receive the previous 
client's model for training in each training round, and when the clients in the ring topology receive the models from all the clients 
it will use the FedAvg algorithm for model aggregation. Lee et al. proposed a federated learning algorithm based on ring topology 
called TornadoAggregate to improve the model accuracy as well as the scalability of federated learning \cite{lee2020tornadoaggregate}. 
Lu Han et al.\cite{han2023ringffl}. considers that dishonest clients can seriously compromise security and fairness in the training 
of ring topology federated learning. To solve the problem, They propose a Ring-architecture-based Fair Federated Learning framework 
and record the transactions of participating clients into a blockchain to prevent tampering. These above federated learning frameworks 
based on ring topology are mainly considered for communication optimization, scalability, fairness and security.
\subsection{Hierarchical Federated Learning}
    To address the bottleneck of servers in traditional federated learning, several studies have proposed the hierarchical federated 
learning framework. Liu et al.\cite{liu2020client} proposed a basic hierarchical federated learning framework, called HierFAVG, to 
reduce the communication cost and speed up the convergence of the model. HierFedAVG allows edge servers to aggregate partial models 
uploaded from clients and the cloud service periodically aggregates models from edge servers, which speeds up model convergence and 
better balancing of communication computations. To achieve low-latency and energy-efficient federated learning, HFEL\cite{luo2020hfel} 
decomposes the client-edge-cloud hierarchical federated learning into two subproblems: resource allocation and edge association. By 
employing the optimal policy for the convex resource allocation subproblem among a group of devices connected to a single edge server, 
HFEL can attain an effective edge association strategy through an iterative process that reduces global costs. To enable continuous 
evolution of the distributed model with new data from devices, Zhong et al.\cite{zhong2022flee} proposed FLEE, a hierarchical federated 
learning framework. FLEE consides various data distributions on devices and edges and allows for dynamic model updates without redeployment. 
TiFL\cite{chai2020tifl} employs an adaptive tier selection approach to mitigate the impact of resource and data heterogeneity by observing 
training efficiency and accuracy to divide clients to different tiers (e.g., fast, and slow tiers) in real time. However, TiFL's strategy 
of favoring faster tiers may lead to training bias and lower model accuracy. FedAT \cite{chai2020fedat} draws on the TiFL tier selection 
approach, but different from TiFL, FedAT combines intra-tier synchronous training with inter-tier asynchronous training in order to 
effectively avoid training biases.
\par
    Different from previous hierarchical federated learning approaches, we focus on solving the problem of data heterogeneity in 
cloud-edge-device hierarchical federated learning. We design a semi-decentralized federated learning framework by combining DFL with 
hierarchical federated learning. In our approach, each device will select a nearby edge server to participate in the training, and 
the edge server organizes these participating training devices into a ring cluster. In each ring cluster we use an incremental 
optimization algorithm for training to mitigate the effect of data heterogeneity. Finally, the cloud server will employed FedAvg 
algorithm to aggregate all edge server models. In the following sections, we will elaborate on the algorithm proposed in this paper.
    \section{Methodology}
    \subsection{Problem Statement}
        In this paper, we consider a federated learning problem that devices collaboratively train a global model by sharing model parameters 
    with the assistance of edge servers and a cloud server. We assume that there are $K$ devices denoted as $P = \{P_1, P_2, ..., P_K\}$. There 
    are M edge server and the $m-th$ edge server is connected by devices in $I_m =  \{P_{m1}, P_{m2}, \cdots \}\subset P$. A device must connect 
    to one edge server and cannot connect to more than one edge server simultaneously. Each device $P_{mi}$ owns its private dataset $D_{mi}$ and 
    $ D_m=\cup_{i=1}^{|I_m|}D_{mi} $ is the set of all the data owned by its connected devices. The set of all the data involved in training a 
    task model with parameter $\mathbf{w}$ can be denoted by $ D = \cup_{m=1}^M D_m$. The loss function for learning the task model is
    \begin{align}
    \label{eq:FL_loss}
    L(\mathbf{w}) &= \frac{1}{|D|}\sum_{(x,y)\in D} l(x,y;\mathbf{w}) \notag \\
    &= \sum_{P_{mi}\in P}\frac{|D_{mi}|}{|D|}\left[\frac{1}{D_{mi}}\sum_{(x,y)\in D_{mi}} l(x,y;\mathbf{w})\right] \notag \\
    &=\sum_{P_{mi}\in P}\frac{|D_{mi}|}{|D|} L_{mi}(\mathbf{w}) 
    \end{align}
    where \[ L_{mi}(\mathbf{w}) = \frac{1}{|D_{mi}|}\sum_{(x,y)\in D_{mi}} l(x,y;\mathbf{w})\] is the loss function of the local model of 
    client $P_{mi}$ on its private dataset $\mathcal{D}_{mi}$. Our goal is to solve a $\mathbf{w}$ minimizing loss function $L(\mathbf{w})$.

    \subsection{Motivation}
        In centralized learning, we typically divide the dataset into multiple batch data and utilize these batch data to compute the 
    gradients of the model parameters for updating it, as follow:
    \begin{equation}
        w = w-\eta \nabla l(w; b)
    \end{equation}
    where $l(w; b)$ represents the loss of $w$ in batch data $b \in [D]$ and the $D$ represents the whole dataset. Similarly, we can 
    consider a device participating in federated learning as a "batch data" in centralized learning. In a general FL system, each "batch 
    data" separately calculates a gradient to update the local model, and a cloud server will aggregate the local model trained by all 
    "batch data", as follow:
    \begin{equation}
        w = w-\eta \sum_{i=1}^K \frac{\left|D_i\right|}{|D|} \nabla l_{i}(w; b_i)
    \end{equation}
    where $l_{i}(w; b)$ represents the loss of $w$ in batch data $b \in [D_i]$ and the $D_i$ represents the local dataset of device $P_i$.
    However, previous research \cite{karimireddy2020scaffold} has shown that when the data distribution across devices is non-iid, it will 
    degrade the performance of the global model. Because in the non-iid scenario, the models trained by each device will be biased towards 
    the local optimum, making the aggregated global model differ significantly from the true optimum. \cite{bertsekas2011incremental} proposed 
    an incremental subgradient optimization algorithm, which consider an optimization problems with a cost function consisting of a lager number 
    of component functions such as
    \begin{equation}
        \label{incremental}
        \min \sum_{i=1}^K f_i(x)
    \end{equation}
    where $f_i$ is real-valued function of $x$. We will use the incremental subgradient optimization algorithm to solve in a cyclic way, as follows:
    \begin{equation}
        x_{k+1}=x_k-\alpha_k \nabla f_{k}\left(x_k\right)
    \end{equation}
    where $\alpha_k$ is a positive stepsize. Through the incremental subgradient optimization algorithm, it can acquire an optimal solution $x^*$ satisfing 
    the equation (\ref{incremental}). Motivated by the incremental subgradient optimization algorithm, we connect devices participating in federated learning 
    in a ring topology, where each device can communicate with neighboring devices. We treat the data of each device as a centralized learning's "batch data". 
    And each device receives the model optimized by the previous device and uses own private data for training, as follows:\\
    1) for clients $k=P_1,P_2,\cdots,P_K$ do
    \begin{align}
        \mathbf{z}_t^{k}=\left\{
            \begin{array}{ll}
                \mathbf{w}_t - \eta_t\bigtriangledown L_k(\mathbf{w}_t; b_k), & k=P_1\\
                \mathbf{z}_t^{k-1} - \eta\bigtriangledown L_k(\mathbf{z}_t^{k-1}; b_k), & k=P_2,\cdots,P_K\\
            \end{array}
        \right.
    \end{align}
    where $l_{i}(w_{i-1}; b)$ represents the loss of $w_{i-1}$ in batch data $b \in [D_i]$ and the $D_i$ represents the local dataset of device $P_i$.\\
    2) update 
    \begin{align}
        \mathbf{w}_{t+1} = \mathbf{z}_t^{P_K}
    \end{align}
    3) repeat the above iteration until stop creteria is met.
    \par
\begin{figure}[h]
        \centering
        \includegraphics[width=0.45\textwidth]{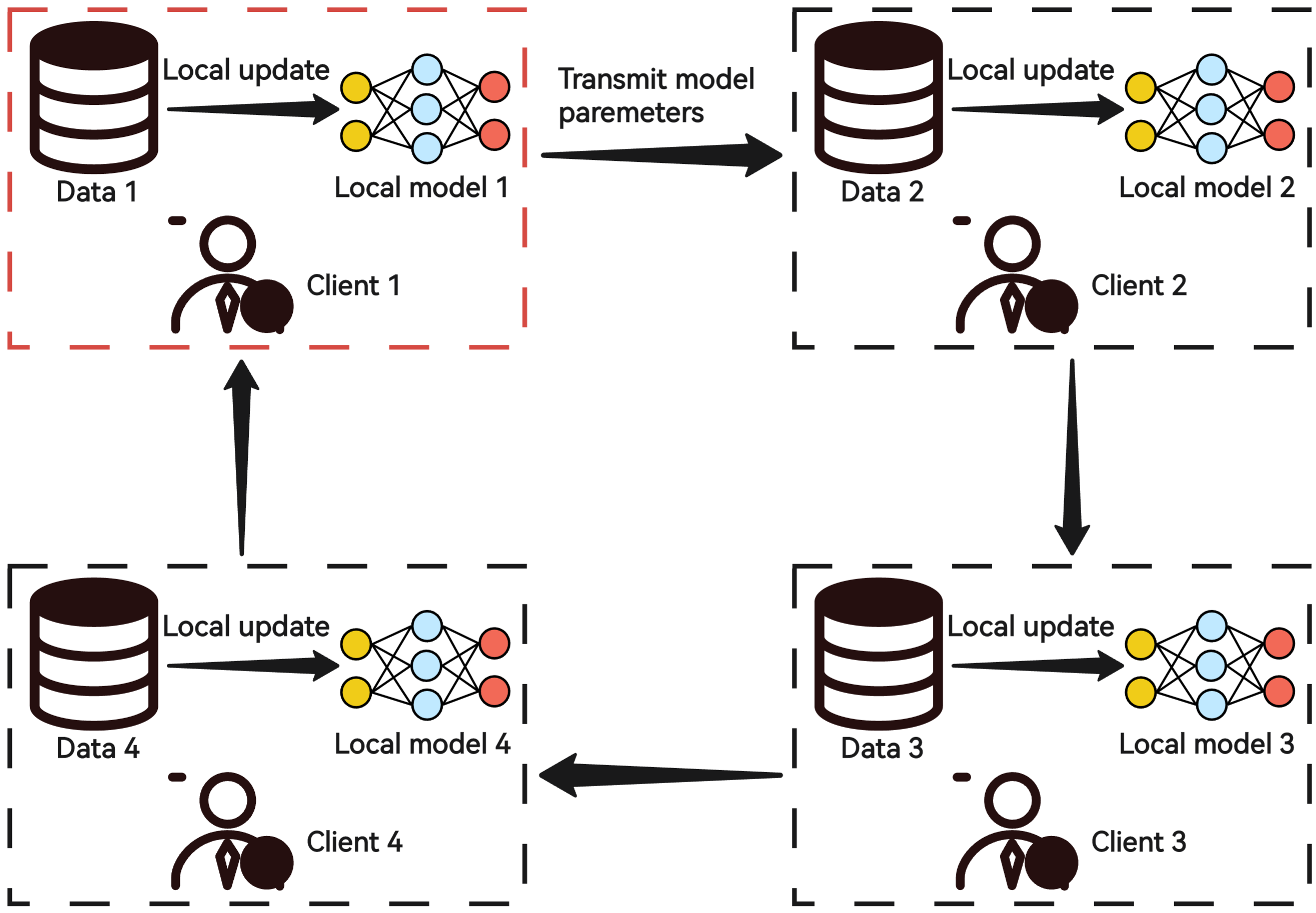}
        \captionsetup{justification=justified}
        \caption{Schematic of a single round of the ring-optimization method, where three steps are implemented sequentially.
                (a) Step1: device 1 receive the model from device 4 to train and update it based on its own dataset.
                (b) Step2: device 1 transmits the locally updated model to device 2.
                (c) Step3: Repeat steps (a) and (b) on the ring topology.
        }
        \label{fig:ring-optimization}
    \end{figure}

    For simplicity, we refer to the above process as ring-optimization and as shown in Fig. \ref{fig:ring-optimization}. Typically the incremental 
    subgradient optimization algorithm is suitable for the case of a large number of devices' participation, however, we experimentally show that 
    a small number of devices' participation also achieves better model performance compared to the traditional federated learning algorithm.
    \par
\begin{figure}[h]
        \centering
        \begin{minipage}[b]{\linewidth}
            \centering
            \subfloat[]{\centering\includegraphics[width=1.7in]{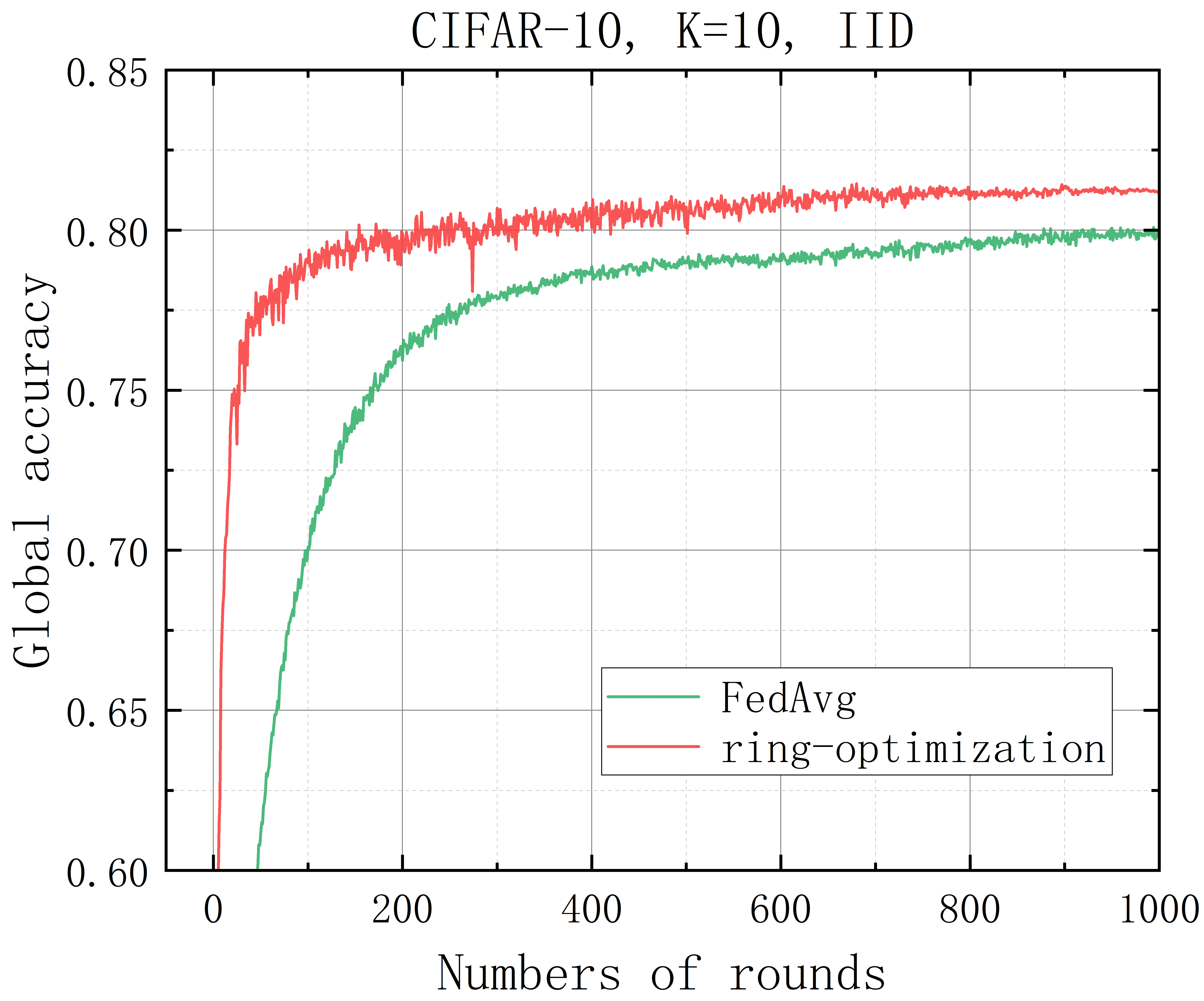}           \label{a}}
            \subfloat[]{\centering\includegraphics[width=1.7in]{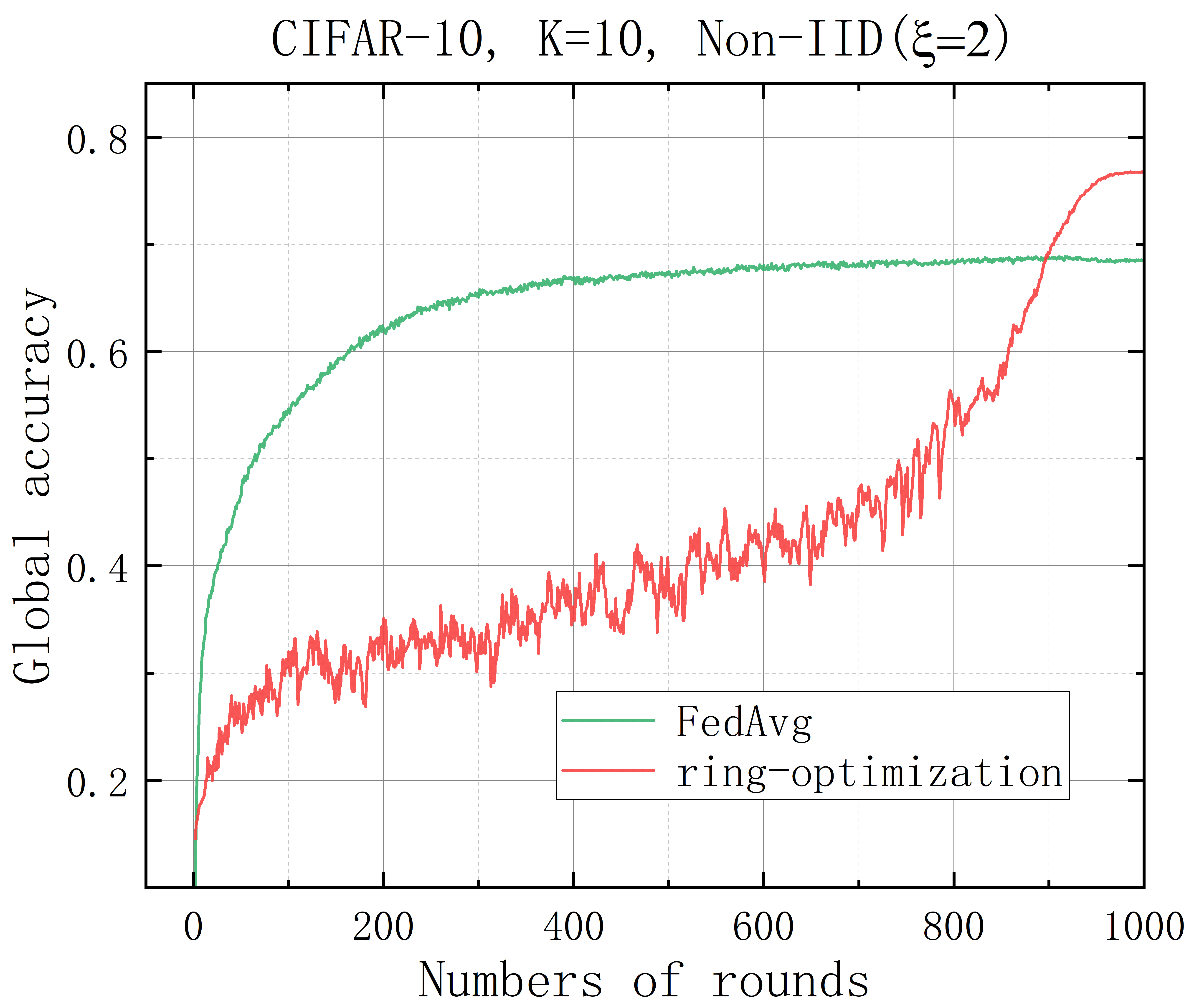}       \label{b}}
        \end{minipage}
        \caption{Classification accuracy on the CIFAR-10 data set using CNN}
        \label{fig:FedRing_IID_and_Non_IID}
    \end{figure}

        To demonstrate the performance of the ring-optimization algorithm when a small number of devices are involved, we conducted experiments on four 
    datasets. In the experiment, the number of devices is set to 10. The neural network model consists of three CNN layers and two MLP layers. We use 
    the SGD optimizer for model updates with a batch size of 32. The initial learning rate is set to 0.01 and decays in a cosine function manner as the 
    training epochs increase. All devices participate in each training round. In this experiment, iid and non-iid setting are considered. For non-iid 
    setting, The CIFAR-100 is divided by pathological distribution ($\xi = 20$) and other datasets are divided by pathological distribution ($\xi = 2$), 
    which can referre to section \ref{sec:Implementation}. We choose FedAvg as the baseline and the number of local update epochs is set to 1 for FedAvg 
    and ring-optimization. And we measured the model prediction accuracy of the global model on the test set. The experimental results are shown in Table 
    \ref{tab:FedRing_IID_and_Non-IID} and Fig. \ref{fig:FedRing_IID_and_Non_IID}.
    \par
    \begin{table}[h]
        \captionsetup{justification=justified}
        \caption{The performance of different methods on different tasks and different data distribution.}
        \resizebox{\linewidth}{!}{
            \begin{tabular}{@{}ccccccc@{}}
                \toprule 
                Task & Method & iid & non-iid($\xi=2$)\\
                \midrule
                \multirow{2}{*}{\rotatebox{0}{\shortstack{MNIST}}}
                & FedAvg               & \textbf{97.66\%}            & 94.38\% \\
                & ring-optimization    & 97.61\%            & \textbf{96.96\%} \\
                \midrule
                \multirow{2}{*}{\rotatebox{0}{\shortstack{FashionMNIST}}}
                & FedAvg               & 92.29\%            & 81.71\% \\
                & ring-optimization    & \textbf{93.09\%}   & \textbf{90.58\%} \\
                \midrule
                \multirow{2}{*}{\rotatebox{0}{\shortstack{CIFAR-10}}}
                & FedAvg               & 80.62\%            & 68.57\% \\
                & ring-optimization    & \textbf{81.20\%}   & \textbf{76.78\%} \\
                \midrule
                Task & Method & iid & non-iid($\xi=20$)\\
                \midrule
                \multirow{2}{*}{\rotatebox{0}{\shortstack{CIFAR-100}}}
                & FedAvg               & 46.05\%            & 40.32\% \\
                & ring-optimization    & \textbf{48.79\%}   & \textbf{44.36\%} \\
                \bottomrule
            \end{tabular}
        }
        \label{tab:FedRing_IID_and_Non-IID}
    \end{table}
        We can find from Table \ref{tab:FedRing_IID_and_Non-IID} that the ring-optimization in both iid and non-iid settings achieves the best global test 
    accuracy in most cases, which demonstrate ring-optimization can effectively address data heterogeneity. Since MNIST is easy to train, ring-optimization 
    and FedAvg achieve high test accuracy in the iid setting. From Fig. \ref{fig:FedRing_IID_and_Non_IID}, We can observe that ring-optimization has a low 
    test accuracy in the early stages of training, however, the test accuracy in the later stages of training surpasses that of FedAvg. We consider that the 
    ring incremental optimization algorithm increases knowledge sharing among devices when performing model optimization. As the learning rate decays, devices 
    can learn the knowledge of other devices in a more detailed way, which enhances the generalization of the local model and avoids biasing the local optimal 
    solution.
    \par
    \subsection{Framework}
    \begin{figure*}[htbp]
        \centering
        \includegraphics[width=1\textwidth]{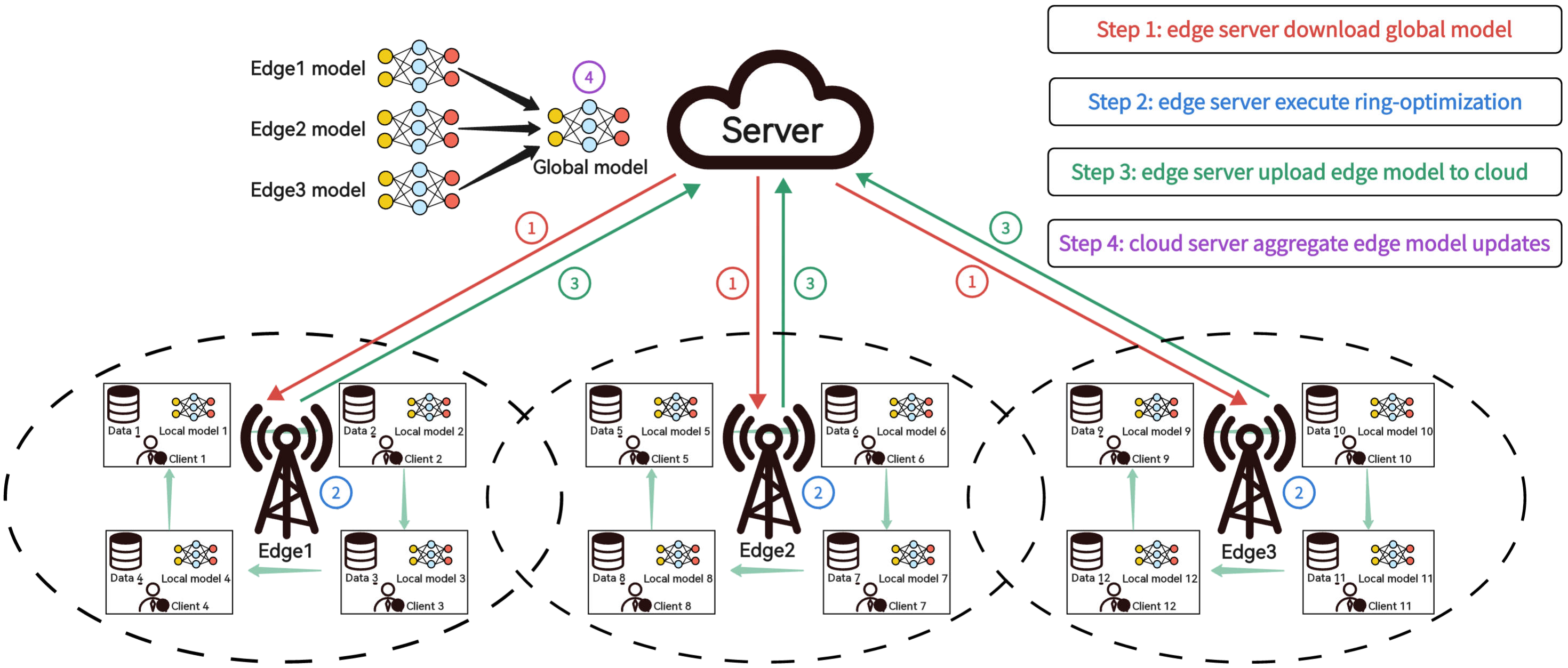}
        \caption{FedSR framework}
        \label{fig:5}
    \end{figure*}
        Although ring-optimization effectively addresses the data heterogeneity, it is not suitable for large-scale training due 
    to the unacceptable training time required when dealing with a large number of devices and converges slowly. However, we can 
    get an insight from ring-optimization that devices in small-scale ring topology can improve the generalization of local models 
    by sharing local model and training it with small learning rates at later stage. Hence, we further propose a federated learning 
    framework for star-ring topology (FedSR), which can be applied to large-scale and non-iid scenarios. If we rewrite the loss function 
    (\ref{eq:FL_loss}) as:\\
    \begin{align}
        \label{eq:FedSR_loss}
        L(\mathbf{w}) &= \frac{1}{|D|}\sum_{(x,y)\in D} l(x,y;\mathbf{w}) \notag\\
        &= \sum_{m=1}^M \frac{|D_m|}{|D|}\sum_{i=1}^{|I_m|}\frac{|D_{mi}|}{|D_m|}L_{mi}(\mathbf{w}) \notag\\
        &= \sum_{m=1}^M \frac{|D_m|}{|D|} L_{m}(\mathbf{w})
    \end{align}
    where \[ L_{m}(\mathbf{w}) = \sum_{i=1}^{|I_m|}\frac{|D_{mi}|}{|D_m|}L_{mi}(\mathbf{w})\]
    The above equation suggests that if we regard $L(\mathbf{w})$ as the loss function of a virtual task model $\mathbf{w}^m$ 
    with loss function $L_{m}(\mathbf{w})$ on edge server $m$ and $D_m$ as its local dataset, then FedAvg can be applied to 
    minimize $L(\mathbf{w})$. And further, the update of $\mathbf{w}^m$( we call it edge model for convenience ) can be made 
    in the cyclic way as the incremental optimization algorithm.\\

    \begin{algorithm}[htbp]
        \caption{
        FedSR. 
        The K devices are indexed by k and M Edge-Servers are indexed by m; 
        E is the number of local epochs, 
        R is the round of FedRing and $\eta_t$ is the learning rate.}
        \textbf{Server executes}:\\
        {initialized $\mathbf{w}_{glob}^{0}$}.\\
        \For{each round t=0,1,...}{
            \For{each $\textbf{Edge-Server}_m$ m=1,2,...M}{
                $\textbf{Edge-Server}_m$ randomly connects devices into a ring network topology.\\
                $\mathcal{I}_{t}^{m}=\{P_{m1}^{t},P_{m2}^{t}...P_{mQ}^{t}\}$, $m \in \{1, 2...M\}$\\
                $w_{t}^{m}$ $\leftarrow$ ring-optimization($\mathcal{I}_{t}^{m}$, $w^t_{glob}$, $\eta_t$, R)
            }
            server aggregate the model from $\textbf{Edge-Server}_m$:
            $\mathbf{w}_{glob}^{t+1} = \sum_{m=1}^M \frac{\left|\mathcal{D}_m\right|}{|\mathcal{D}|} \mathbf{w}_t^{m}$
        }
        \textbf{ring-optimization}($(\alpha_1, \alpha_2, ...\alpha_K)$, $w_{glob}$, $\eta$, R):\\
        $w_{0}^{\alpha_1}$ $\leftarrow$ $w_{glob}$\\
        \For{each round r=0,1,...R}{
            \For{k=1,2,...K}{
                \eIf(){$k \neq K$}{
                    $w_{r}^{\alpha_{k+1}}$ $\leftarrow$ $\mathrm{SGD}\left(\mathbf{w}_r^{\alpha_{k}}, \mathcal{D}_{\alpha_{k}}\right)$
                }(){
                    $w_{r}^{\alpha_{1}}$ $\leftarrow$ $\mathrm{SGD}\left(\mathbf{w}_r^{\alpha_{k}}, \mathcal{D}_{\alpha_{k}}\right)$
                }
            }
        }
        return $w_{R}^{\alpha_{K}}$
        \label{algorithm:3}
    \end{algorithm}
        The framework of FedSR is shown in Fig. \ref{fig:5}. In each training round of FedSR, we divided it into ring-optimization 
    phase and model aggregation phase. In the ring optimization phase, we employ the incremental subgradient optimization algorithm 
    to train a edge server model. Firstly, the cloud server broadcasts the global model $w_{glob}^{t}$ to all edge server. On edge 
    server $m$, the first device $P_{m1}^{t}$ of each ring network topology will receive a global model $w_{glob}^{t}$ from the edge 
    server and optimize the global model with own private dataset $\mathcal{D}_{P_{m1}^t}$. When device $P_{m1}^{t}$ finishes the model 
    update, it sends its local model to the next device $P_{m2}^{t}$ in the ring network topology. After that, the model training for 
    each device in the ring network topology can represent as follows:
    \begin{align}
        \label{eq:ring}
        \mathbf{z}_t^{mi}=\left\{
            \begin{array}{ll}
            \mathbf{w}_t^m-\eta_t\bigtriangledown g_{mi}(\mathbf{w}_t), &i=1\\
            \mathbf{z}_t^{m,i-1} - \eta_t\bigtriangledown g_{mi}(\mathbf{z}_t^{m,i-1}), &i=2,\cdots,|I_m|
            \end{array}
        \right.
    \end{align}
    and 
    \begin{align}
        \mathbf{w}_{t+1}^m = \mathbf{z}_t^{mK}
    \end{align} 
    In the model aggregation phase, the cloud server will receive the model $\mathbf{w}_t^{k^{t}_{mQ}}$ form edge server, and perform an 
    average weighted aggregation of these models based on the number of datasets of devices in the edge server, as follow:
    \begin{align}
        \mathbf{w}_{glob}^{t+1} = \sum_{m=1}^M \frac{|D_m|}{|D|} \mathbf{w}_{t}^m
    \end{align}
        In FedSR, only models are transferred between devices and each device's private data is reserved locally. To protect the privacy 
    leakage problem between devices, we can also use techniques such as differential privacy and homomorphic encryption. In this paper, 
    we focus on the non-iid problem in federated learning, so we do not further discuss the privacy problem between devices. The specific 
    implementation of FedSR is shown in algorithm \ref{algorithm:3}.
    \subsection{Convergence Analysis Of FedSR}
        To analyze the convergence of the FedSR, we use the assumption below. {\bf Assumption 1:} There is a constant $c \in \mathcal{R}$ such that 
    for all $m$ and $i$
    \begin{align}
        max\left\{\Vert \bigtriangledown f_{mi}(\mathbf{z}_t^{mi})\Vert\right\}\leq c
    \end{align}
    Futhermore, we have
    \begin{align} 
        max \left\{ g_{mi}(\mathbf{w}_t^m) - g_{mi}(\mathbf{z}_t^{m,i-1})\right\}\leq c\Vert \mathbf{w}_t^m-\mathbf{z}_t^{m,i-1}\Vert,
    \end{align}
    \begin{align}
        g_{mi}(\mathbf{z}_t^{m,i-1}) - g_{mi}(\mathbf{z}_t^{mi})\leq c\Vert \mathbf{z}_t^{m,i-1}-\mathbf{z}_t^{mi}\Vert \\
        \forall m=1,\cdots,M;i=2,\cdots,|I_m| \notag
    \end{align}

    {\bf Assumption 2:} There is a constant $ 0<a<1 $ such that
    \begin{align}
        \min_{m=1}^M \frac{|D_m|}{|D|} \geq a
    \end{align}
    Under the above assumptions, it is easy to show 
    {\bf Lemma:} Let $\mathbf{w}_{glob}^t, \mathbf{w}_t^m, \mathbf{z}_t^{mi}$ be the sequences generated by the proposed method, 
    then for any task model parameter $\mathbf{y}$, we have
    \begin{align}
        \Vert \mathbf{w}_{glob}^{t+1} - \mathbf{y} \Vert ^ 2 &\leq 2\sum_{m=1}^M\frac{|D_m|^2}{|D|^2} \left[\Vert \mathbf{w}_{glob}^t - \mathbf{y} \Vert^2 + \eta_t^2|I_m|^2c^2\right] \notag\\
        &- 4a\eta_t \left(L(\mathbf{w}_{glob}^t) - L(\mathbf{y})\right)
    \end{align}
    
    {\bf Proof:} For each edge server, following the result of proposition 3.1 in \cite{bertsekas2011incremental}, we have 
    \begin{align}
        &\Vert \mathbf{w}_{t+1}^{m} - \mathbf{y} \Vert ^ 2\leq \Vert \mathbf{w}_{glob}^{t} - \mathbf{y} \Vert ^ 2 - \notag\\
        &2\eta_t\left(L_m(\mathbf{w}_{glob}^t) - L_m(\mathbf{y})\right) + \eta_t^2|I_m|^2c^2
    \end{align}
    Since
    \begin{align}
    &\Vert \mathbf{w}_{glob}^{t+1} - \mathbf{y} \Vert^2 = \Vert \sum_{m=1}^M \frac{|D_m|}{|D|} \mathbf{w}_{t+1}^m -  \mathbf{y} \Vert^2 \notag\\
    &\leq 2\sum_{m=1}^M \frac{|D_m|^2}{|D|^2}\Vert \mathbf{w}_{t+1}^m -  \mathbf{y} \Vert^2
    \end{align}
    and
    \begin{align}
    &\sum_{m=1}^M \frac{|D_m|^2}{|D|^2}\left(L_m(\mathbf{w}_{glob}^t) - L_m(\mathbf{y})\right) \geq \notag\\
    &a\sum_{m=1}^M \frac{|D_m|}{|D|}\left(L_m(\mathbf{w}_{glob}^t) - L_m(\mathbf{y})\right)=a\left(L(\mathbf{w}_{glob}^t) - L(\mathbf{y})\right)
    \end{align}
    Combining the above relations together yields the result. {\bf Q.E.D.}
    {\bf Thoerem: } Let $\mathbf{w}_{glob}^t, \mathbf{w}_t^m, \mathbf{z}_t^{mi}$ be the sequences generated by the proposed method 
    and $|E|$ = $\sum_{m=1}^M\frac{|D_m|^2}{|D|^2} \leq \frac{1}{2}$, let the learning rate $\eta_t$ satisfy
    \begin{align}
        \lim_{t\to \infty}\eta_t=0,\space \sum_{t=0}^{\infty}\eta_t=\infty,\space \sum_{t=0}^{\infty}\eta_t^2<\infty
    \end{align}
    Then $\mathbf{w}_{glob}^t$ converges to its optimal value $\mathbf{w}^*$.{\bf Proof:} From the Lemma above and the condition $|E|$ = 
    $\sum_{m=1}^M\frac{|D_m|^2}{|D|^2} \leq \frac{1}{2}$, we have
    \begin{align} 
        \Vert \mathbf{w}_{glob}^{t+1} - \mathbf{w}^* \Vert ^ 2 \leq  \Vert \mathbf{w}_{glob}^t \notag - \mathbf{w}^* \Vert^2 + 2\eta_t^2\sum_{m=1}^M\frac{|D_m|^2}{|D|^2}|I_m|^2c^2 \notag\\
        - 4a\eta_t \left(L(\mathbf{w}_{\text{glob}}^t) - L(\mathbf{w}^*)\right)
    \end{align}
        By the Supermartingale Convergence Theorem, the sequence $\{\Vert \mathbf{w}_{glob}^{t+1} - \mathbf{w}^*  \Vert\}$ converges. Then follow the proof of 
    Proposition 4.3 in \cite{bertsekas2011incremental}, $\mathbf{w}_{glob}^t \to \mathbf{w}^*$. {\bf Q.E.D.}

    \begin{figure*}[htbp]
    \centering
    \begin{minipage}[b]{\linewidth}
        \centering
        \subfloat[]{\centering\includegraphics[height=2in, width = 2.25in]{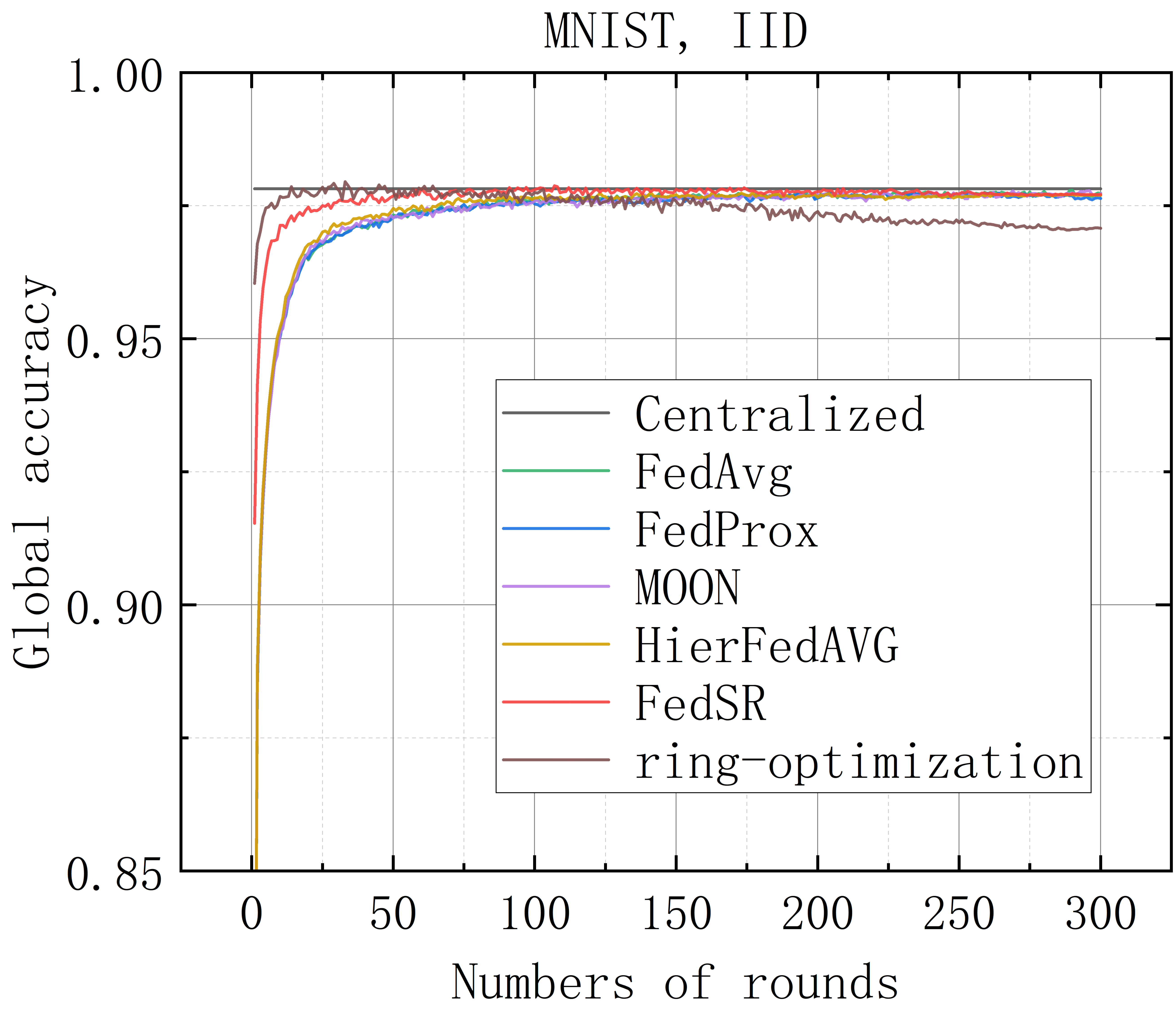}                   \label{a}}
        \hfill
        \subfloat[]{\centering\includegraphics[height=2in, width = 2.25in]{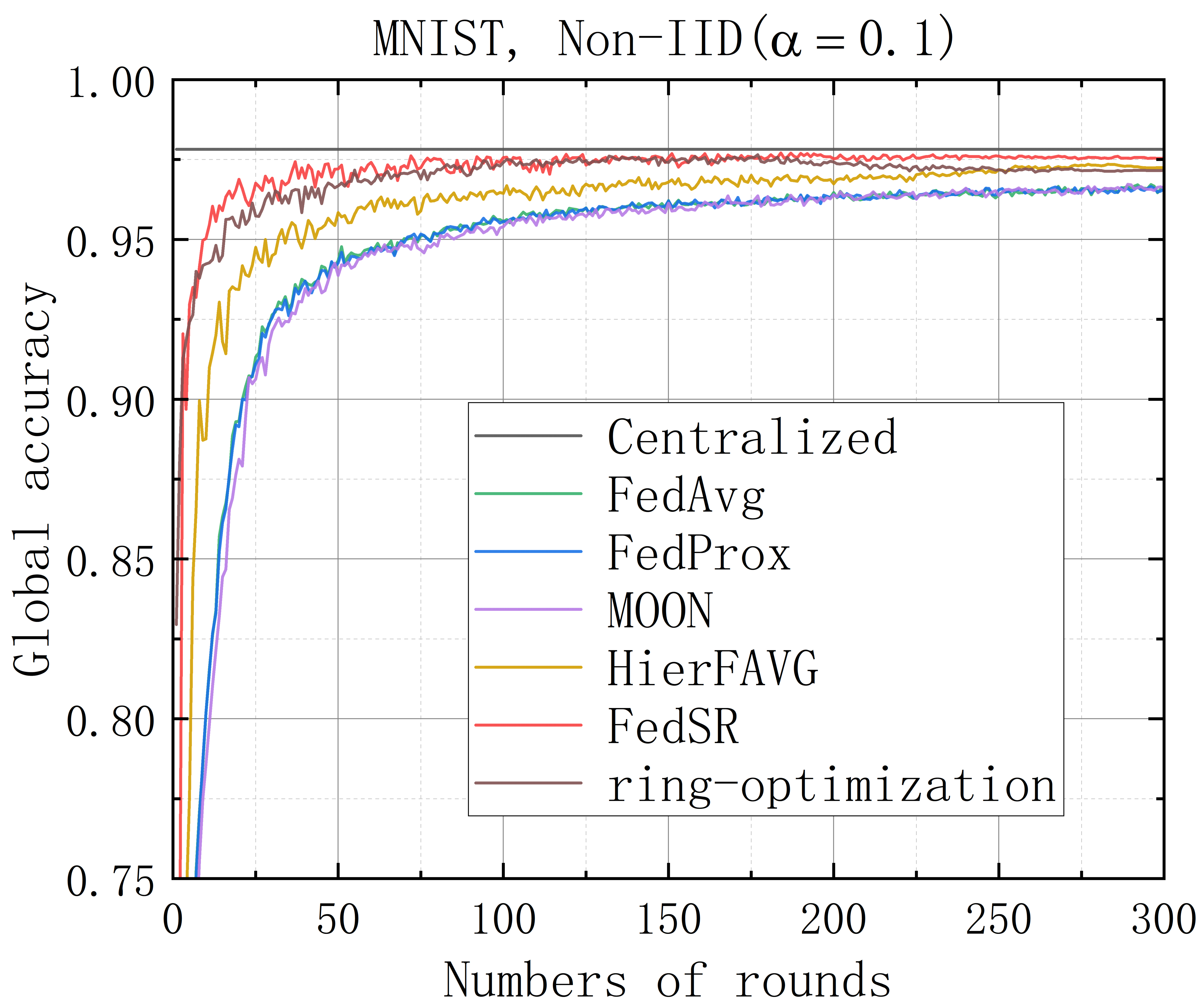}         \label{b}}
        \hfill
        \subfloat[]{\centering\includegraphics[height=2in, width = 2.25in]{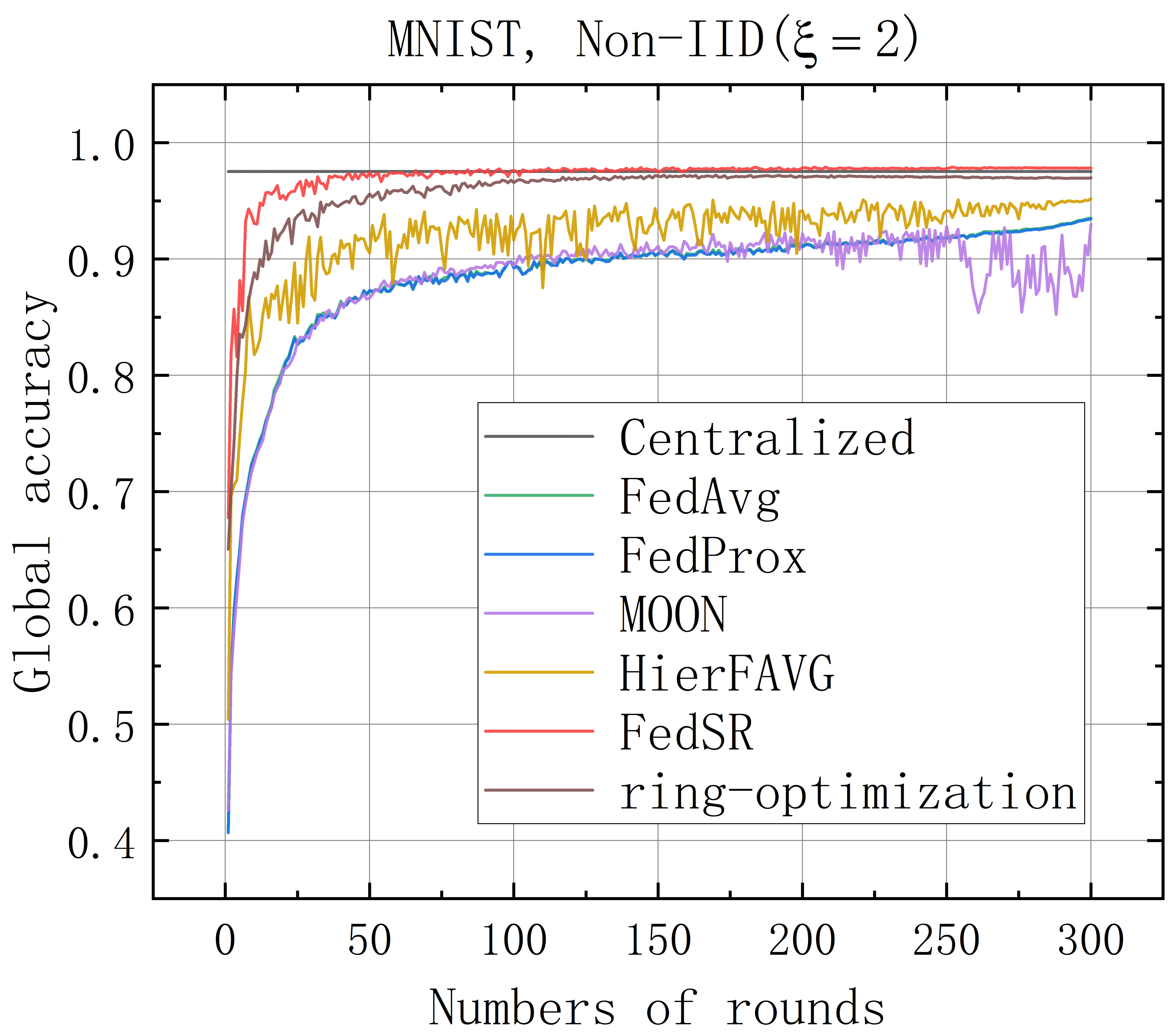}           \label{c}}
        \hfill
        \subfloat[]{\centering\includegraphics[height=2in, width = 2.25in]{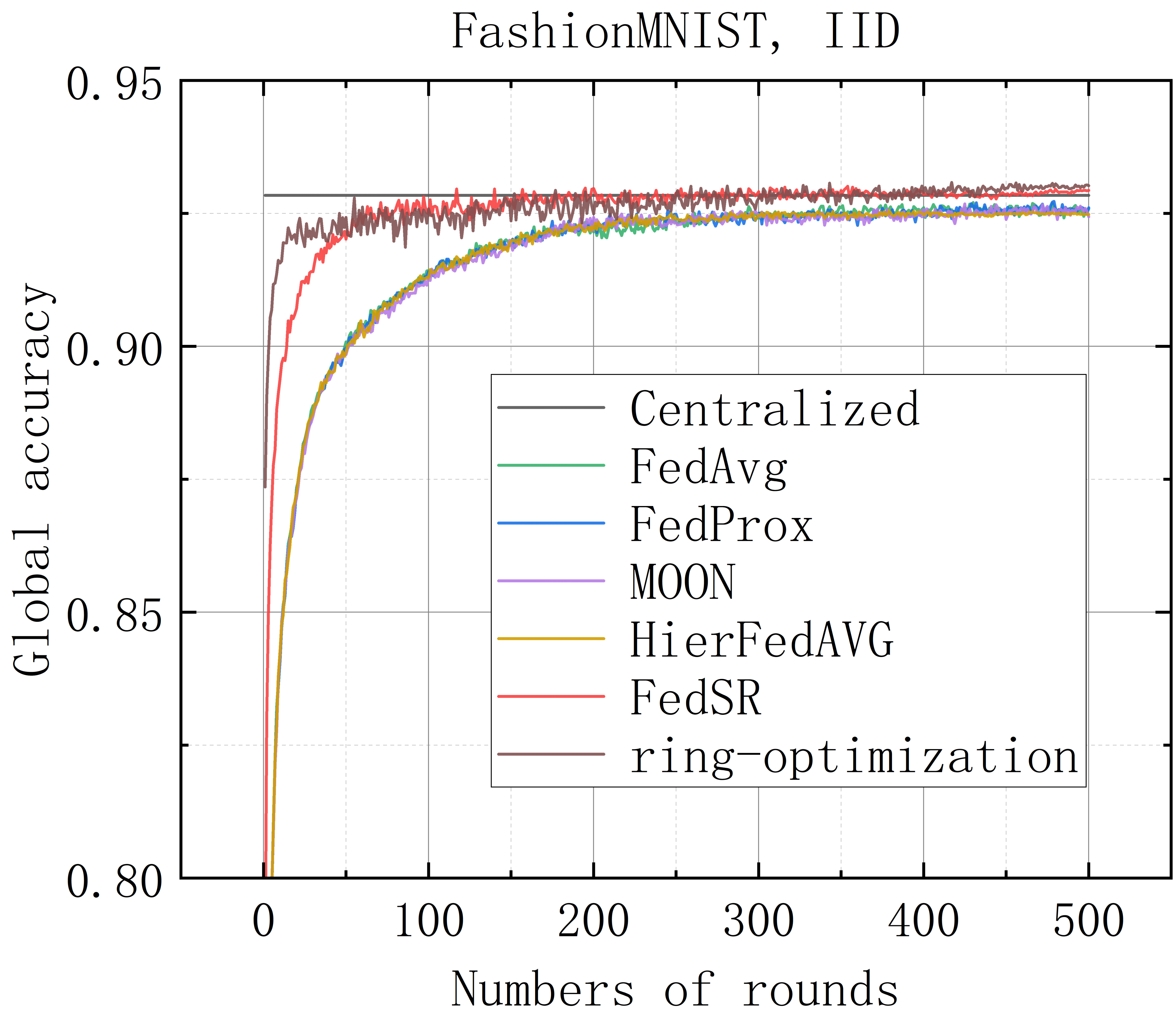}            \label{d}}
        \hfill
        \subfloat[]{\centering\includegraphics[height=2in, width = 2.25in]{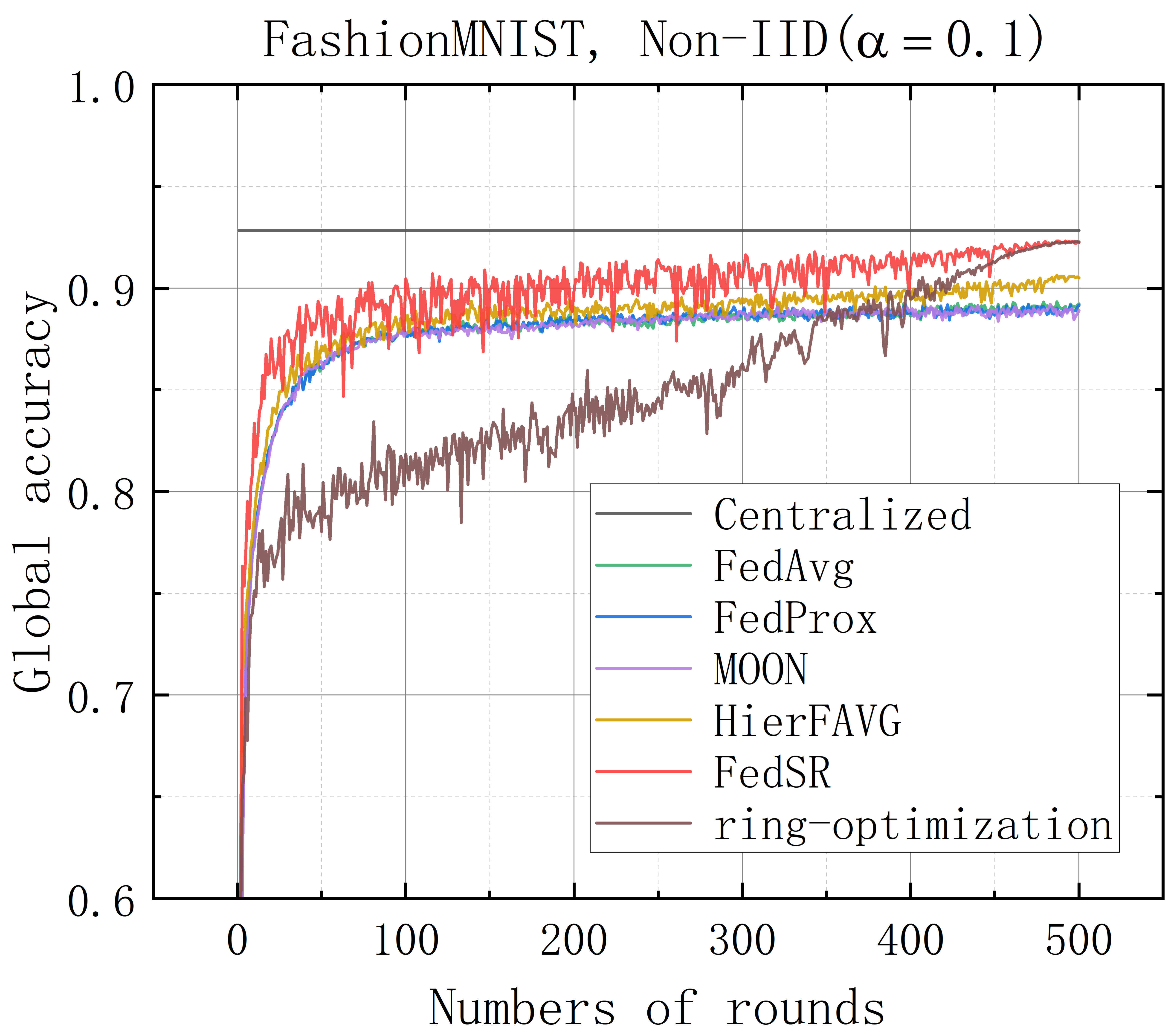}  \label{e}}
        \hfill
        \subfloat[]{\centering\includegraphics[height=2in, width = 2.25in]{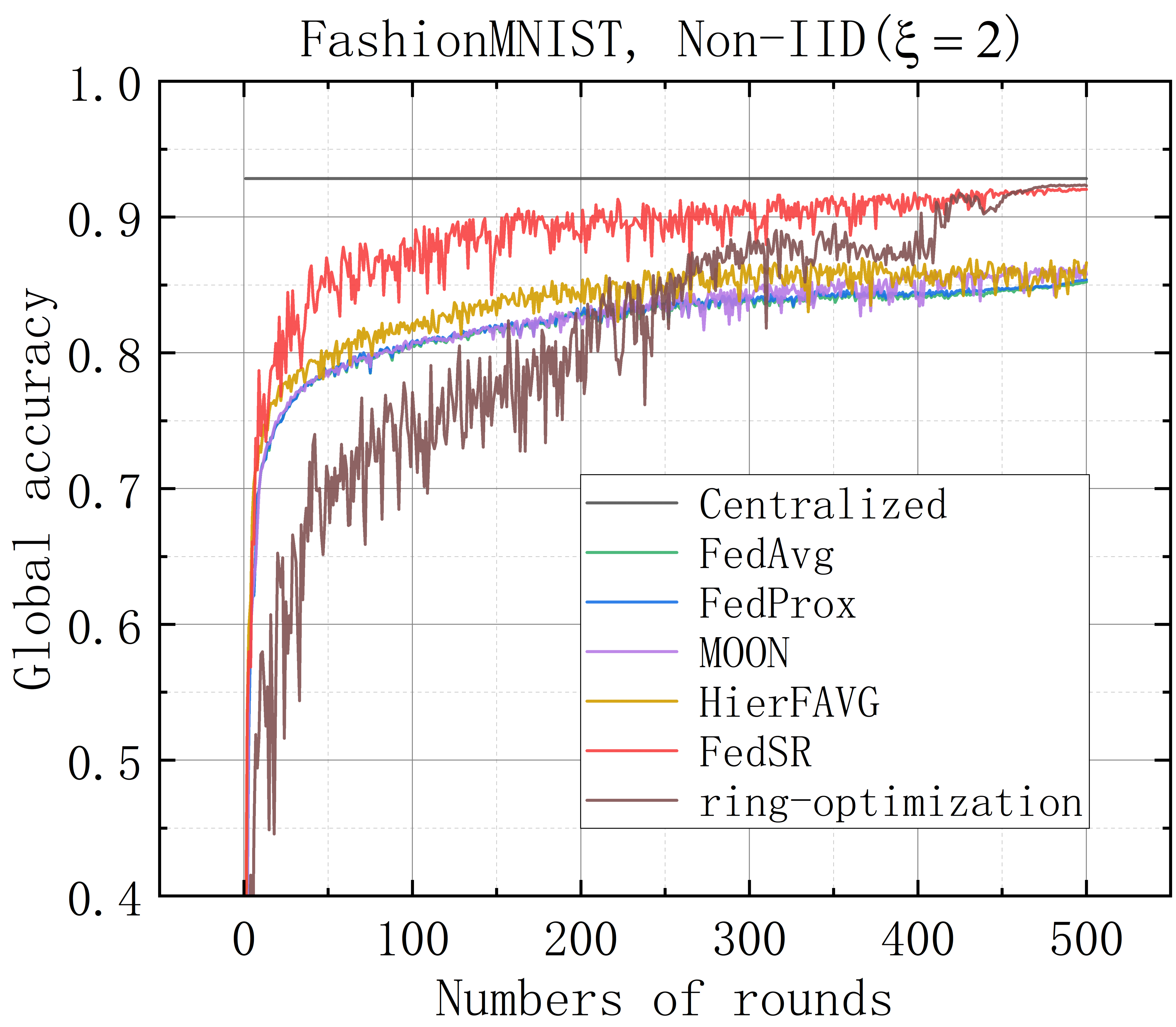}    \label{f}}
        \hfill
        \subfloat[]{\centering\includegraphics[height=2in, width = 2.25in]{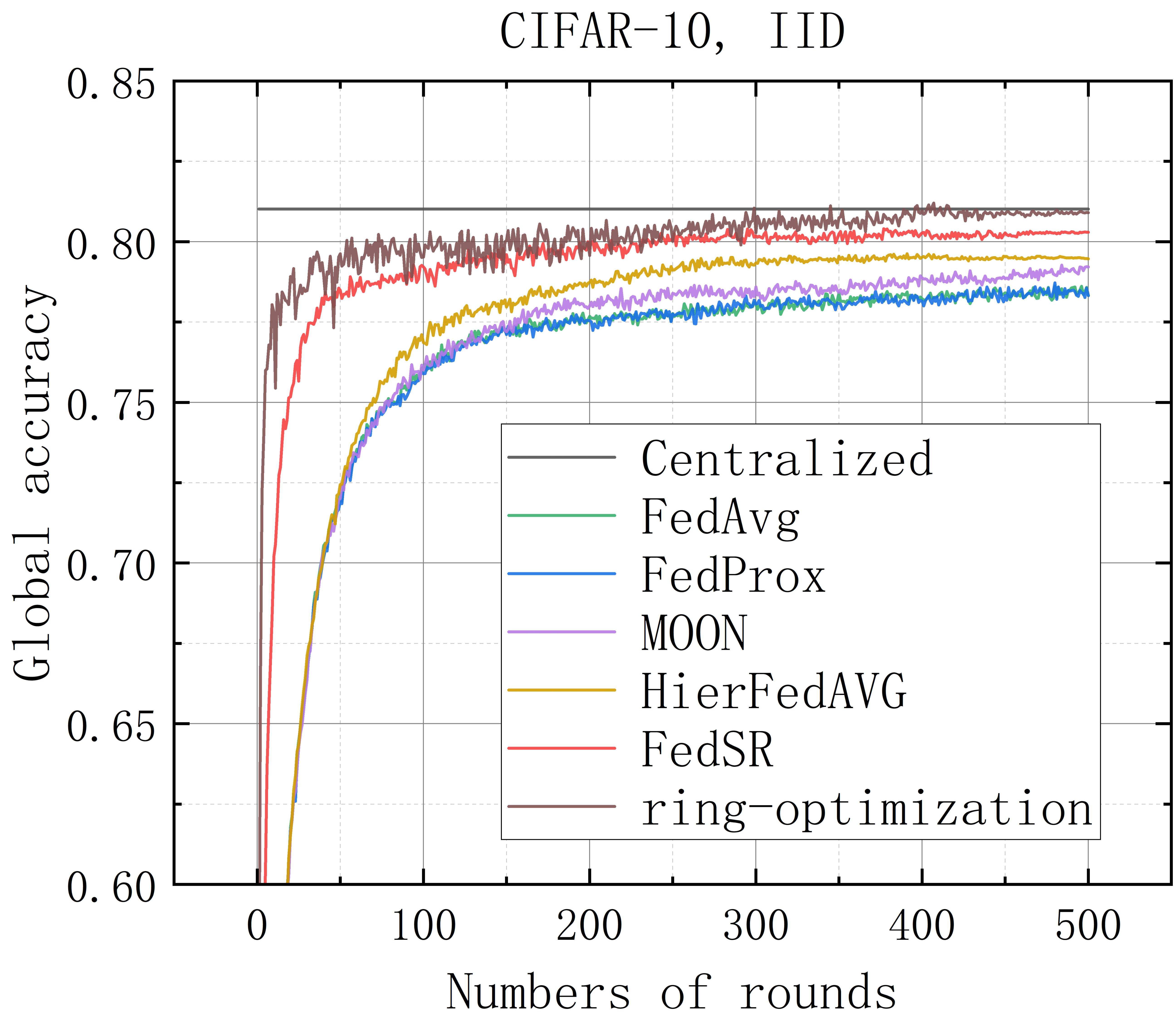}                 \label{g}}
        \hfill
        \subfloat[]{\centering\includegraphics[height=2in, width = 2.25in]{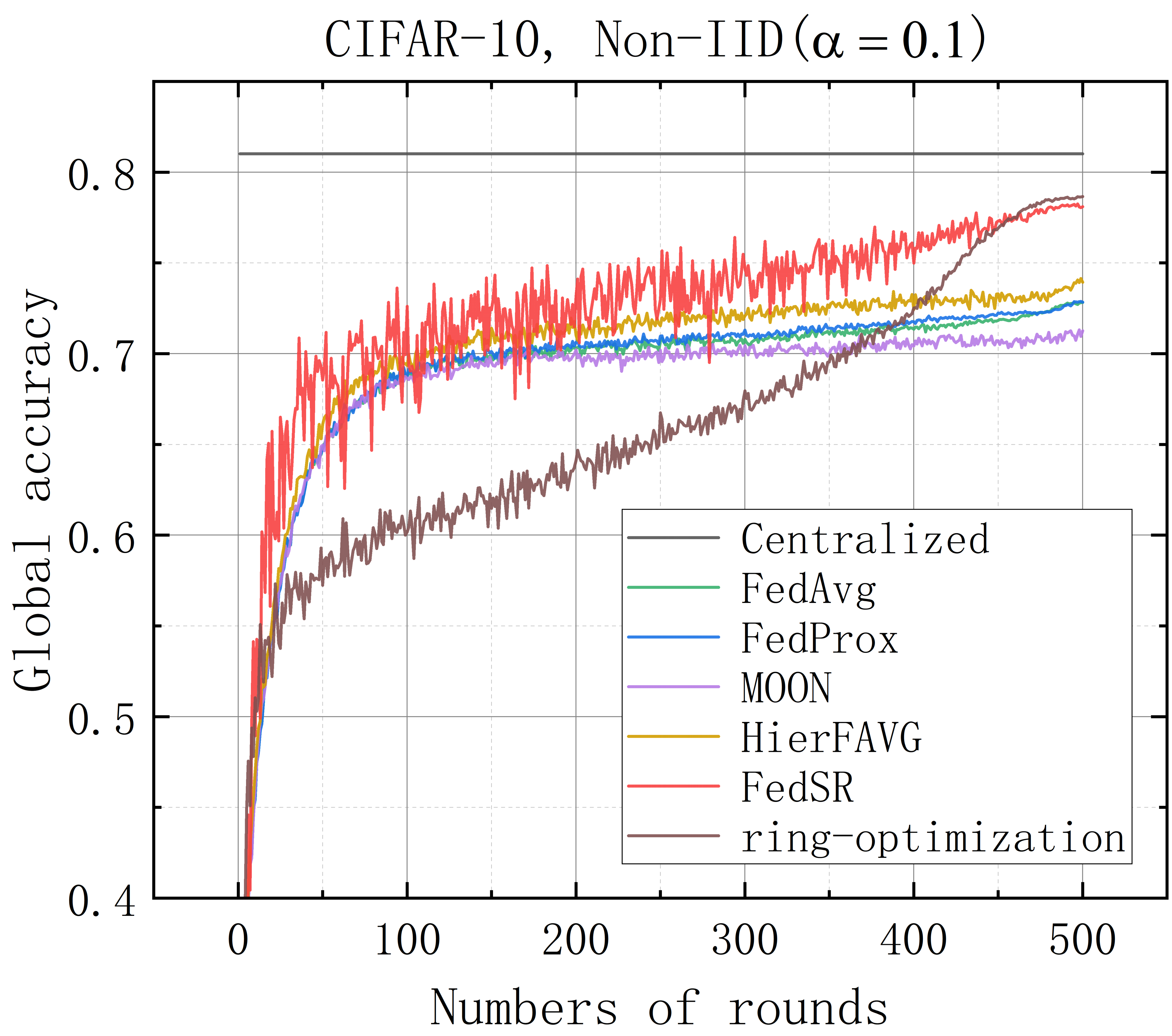}       \label{h}}
        \hfill
        \subfloat[]{\centering\includegraphics[height=2in, width = 2.25in]{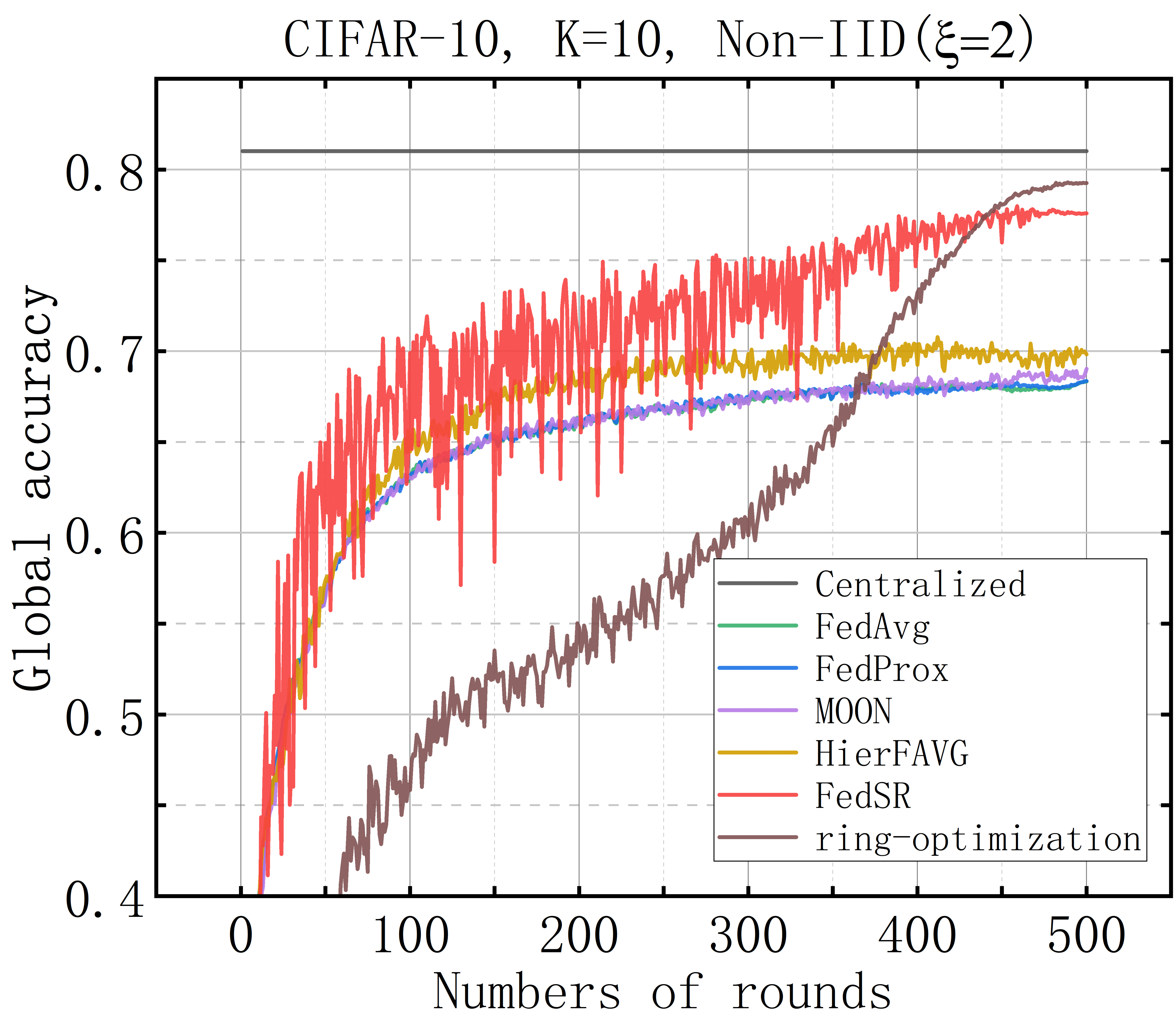}         \label{j}}
        \hfill
        \subfloat[]{\centering\includegraphics[height=2in, width = 2.25in]{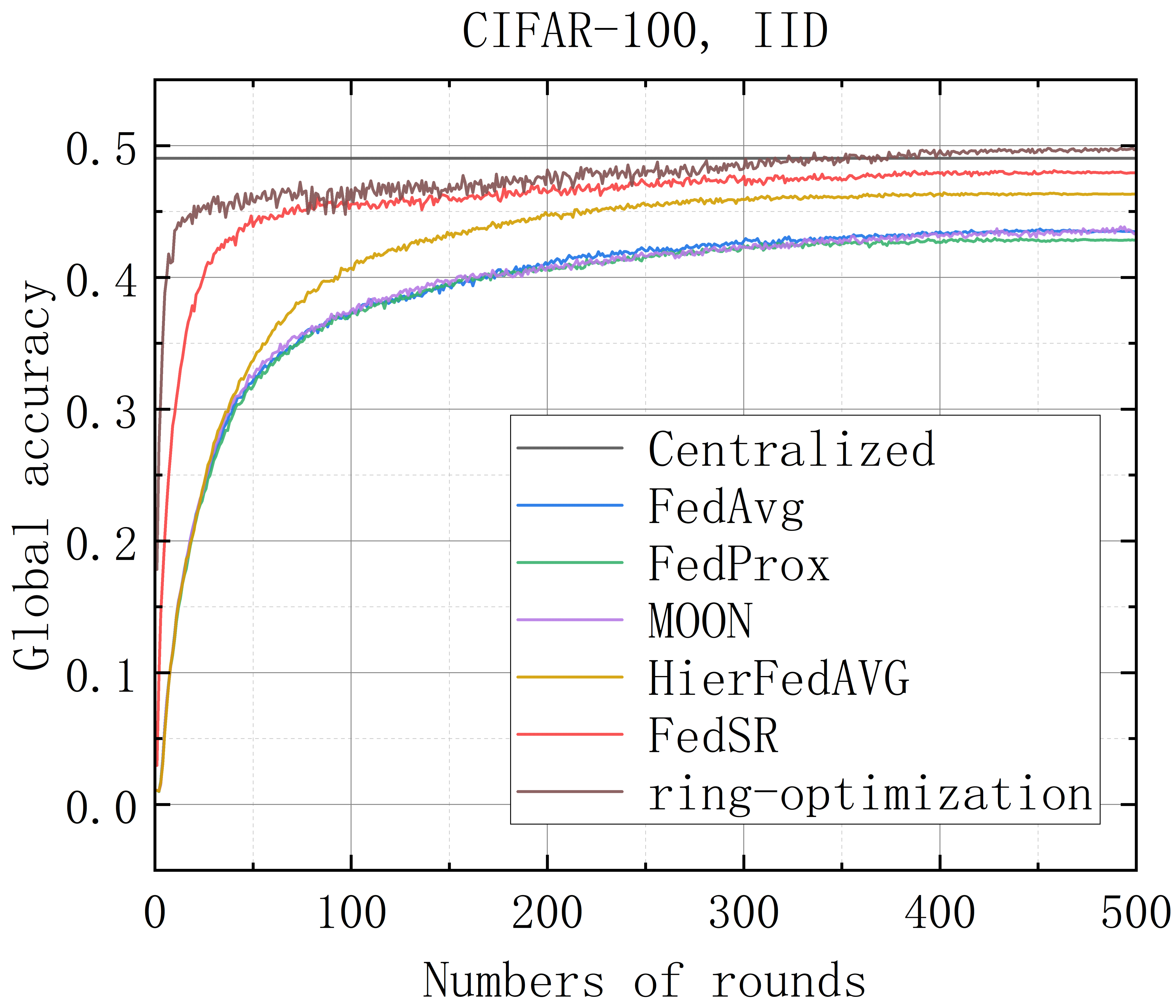}                \label{k}}
        \hfill
        \subfloat[]{\centering\includegraphics[height=2in, width = 2.25in]{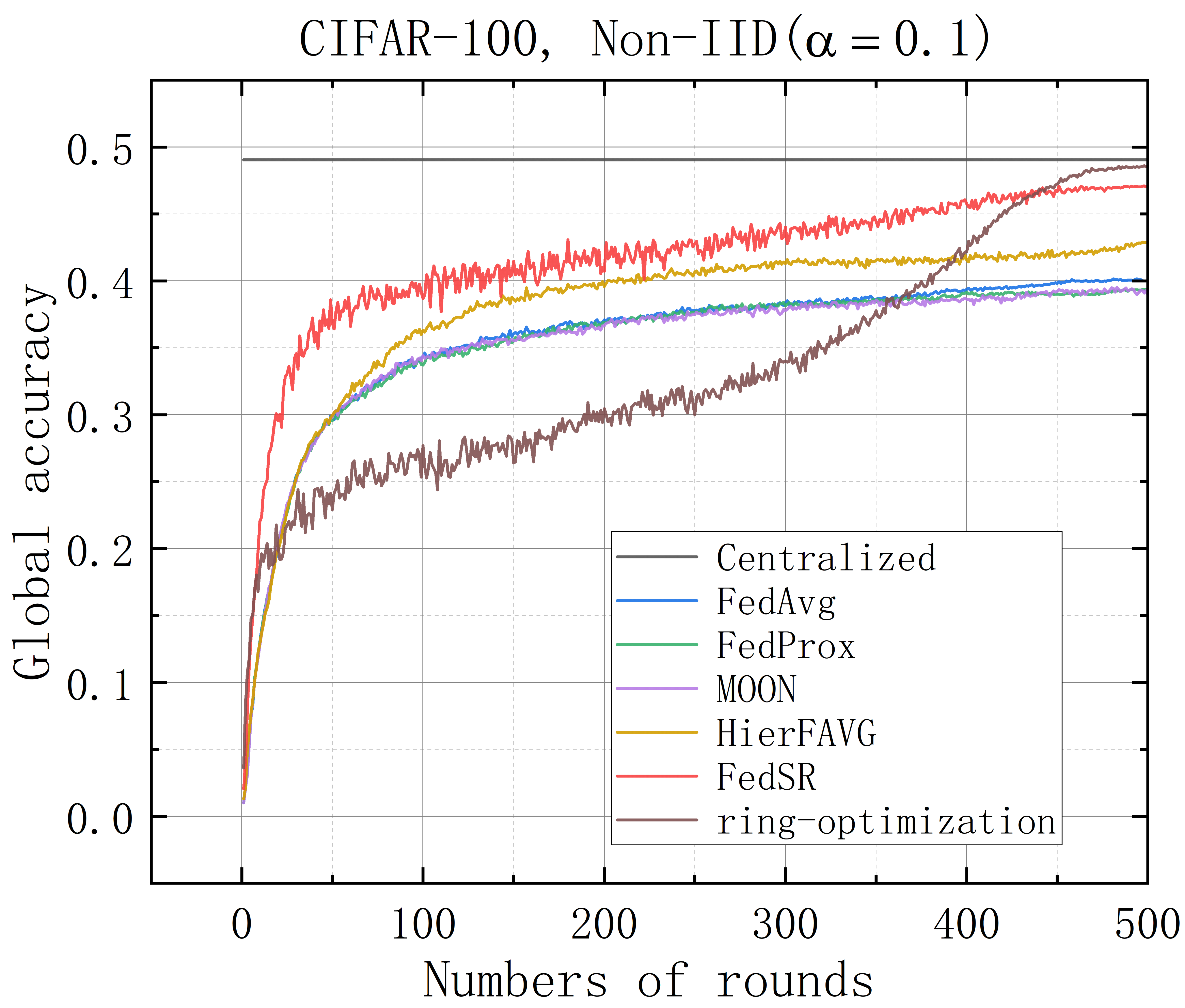}      \label{l}}
        \hfill
        \subfloat[]{\centering\includegraphics[height=2in, width = 2.25in]{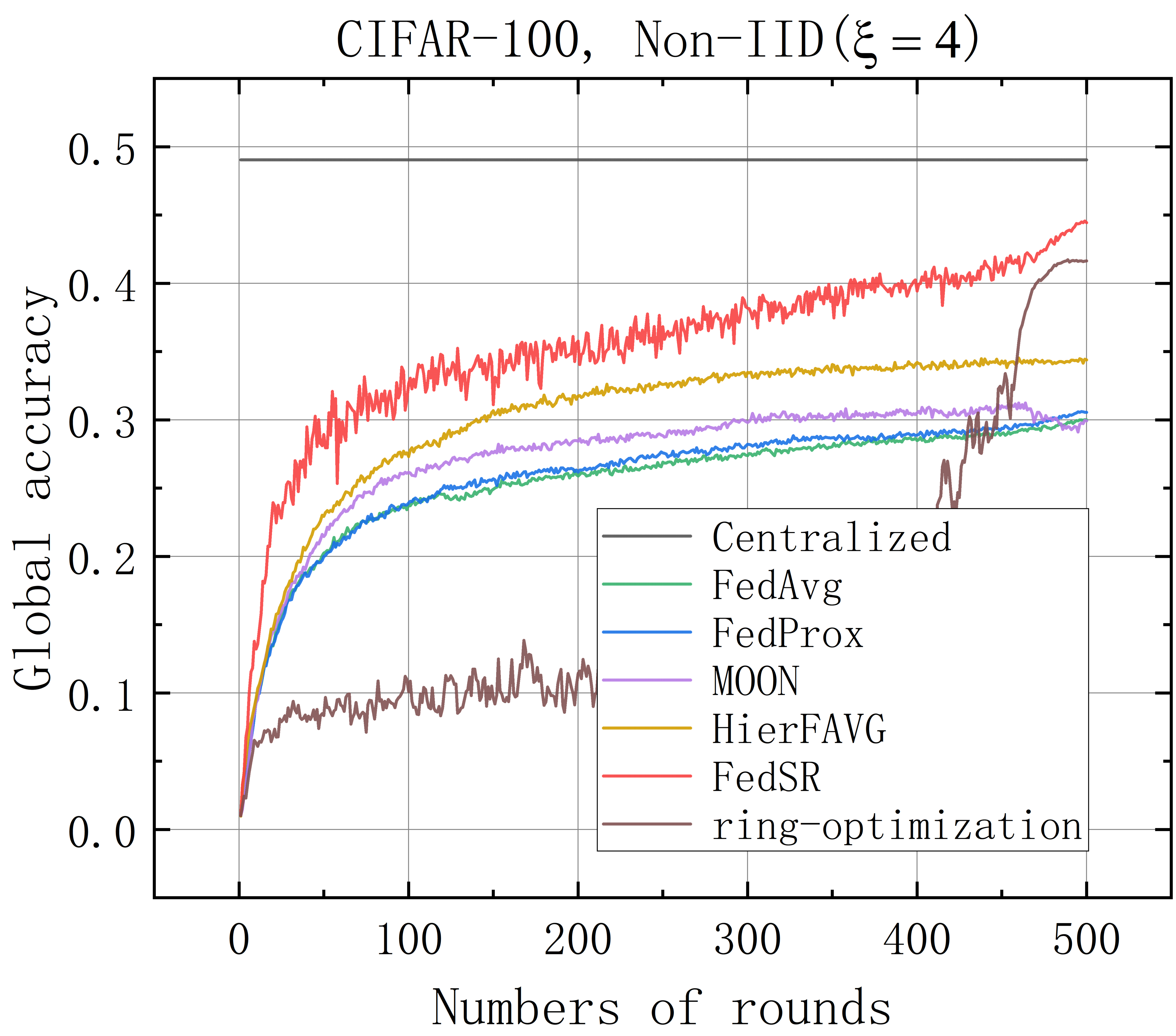}        \label{m}}
    \end{minipage}
    \caption{Classification accuracy on the MNIST, FashionMNIST, CIFAR-10, CIFAR-100 data set, where different algorithms are performed on iid and non-iid setting.}
    \label{fig:IID_and_Non_IID}
\end{figure*}

\section{EXPERIMENTS}
    In this section, we compare the FedSR algorithm with state of the art (SOTA) federated learning algorithms in four image classification tasks to demonstrate that 
our proposed algorithm is able to achieve better model performance in the non-iid case.
\subsection{Data Sets}
    The datasets employed in this experiment are MNIST, FashionMNIST, CIFAR-10 and CIFAR-100. MNIST consists of handwritten digit images from 0 to 9, 
with a size of $28\times28$ pixels, and the training set and test set contain 60k and 10k samples respectively. FashionMNIST consists of 10 classes of 
fashion items with a size of $28\times28$ pixels. The entire dataset has a training set of 60k samples and a test set of 10k samples. CIFAR-10 is an RGB 
image dataset with 10 classes and a size of $32\times32$ pixels. It has a training set and a test set, each containing 50k and 10k samples, respectively. 
CIFAR-100 dataset consists of 100 classes image with a size of $28\times28$ pixels. The training set and test set contain 50k and 10k samples, respectively.
\subsection{Baseline Methods}
    In this experiment, we selected FedAvg, FedProx, MOON, HierFAVG and ring-optimization as baselines. FedAvg periodically broadcasts the global model 
to participating devices for local model training. The server collects these local models trained by devices and aggregates them to form a new global 
model. The FedProx and MOON algorithms enhance the FedAvg approach by respectively incorporating proximal and contrastive losses, with the aim of mitigating 
the effects of data heterogeneity. In FedProx and MOON algorithms, there is a coefficient $\mu$ that constrains the regularization term for local model 
updates. In this experiment, $\mu$ will be carefully adjusted to achieve optimal performance. In HierFAVG, devices can select a nearby edge server to 
participate federated learning. Each edge server aggregates the local models trained by devices using the FedAvg algorithm. The cloud server periodically 
aggregates the models from all edge servers to form the global model.
\subsection{Implementation Details}
\label{sec:Implementation}
\begin{figure}[!h]
    \centering
    \begin{minipage}[b]{\linewidth}
        \centering
        \subfloat[]{\centering\includegraphics[width=1.7in]{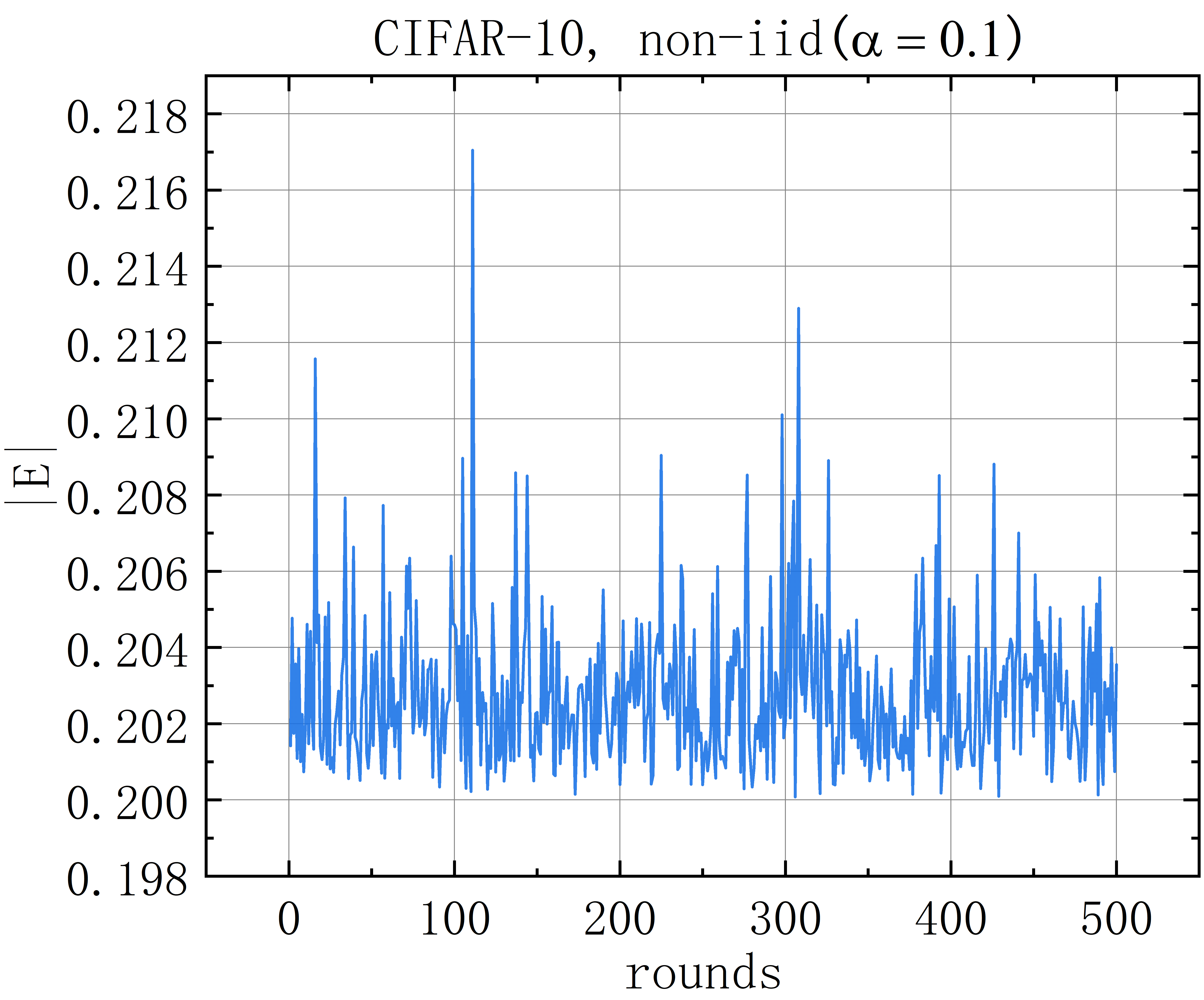}       \label{a}}
        \subfloat[]{\centering\includegraphics[width=1.7in]{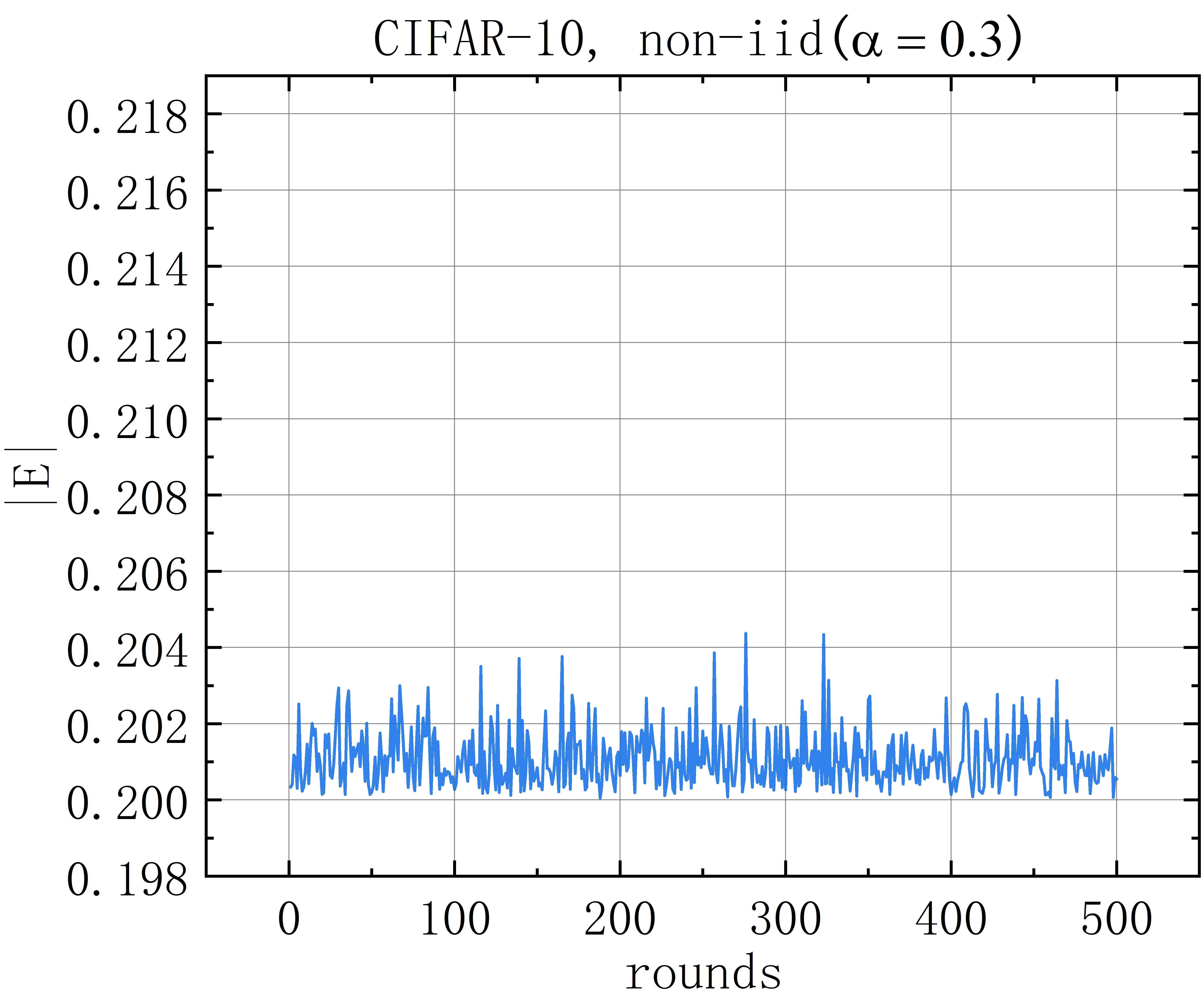}       \label{b}}
    \end{minipage}
    \caption{In the case of the Dirichlet distribution, the sum of squares of weighting coefficients for all edge server model $|E| <= \frac{1}{2}$, satisfying the convergence condition of FedSR}
    \label{fig:weight}
\end{figure}

\begin{figure*}[!h]
    \centering
    \captionsetup{justification=justified}
    \begin{minipage}[b]{\linewidth}
        \centering
        \subfloat[]{\centering\includegraphics[width=2.3in]{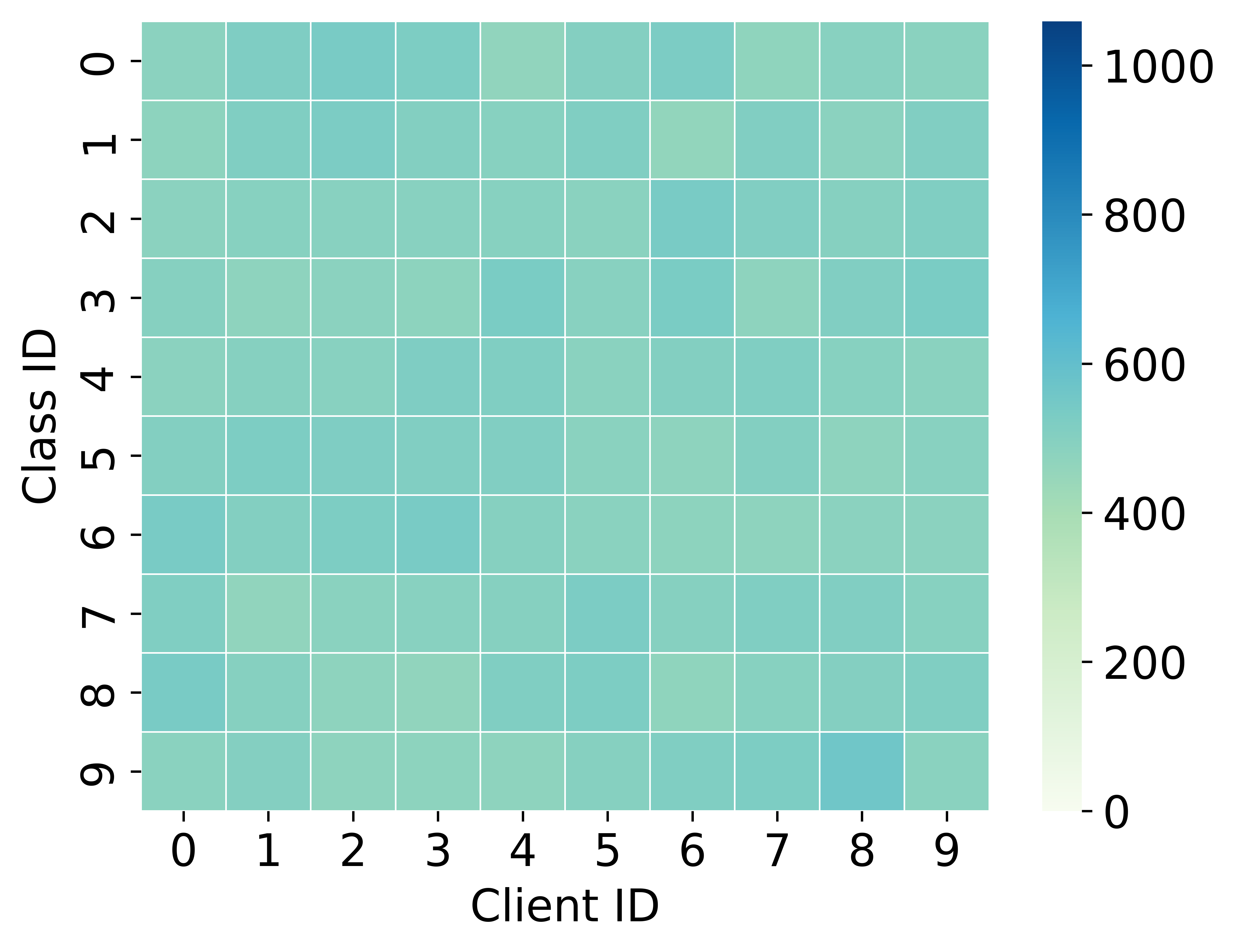}            \label{a}}
        \hfill
        \subfloat[]{\centering\includegraphics[width=2.3in]{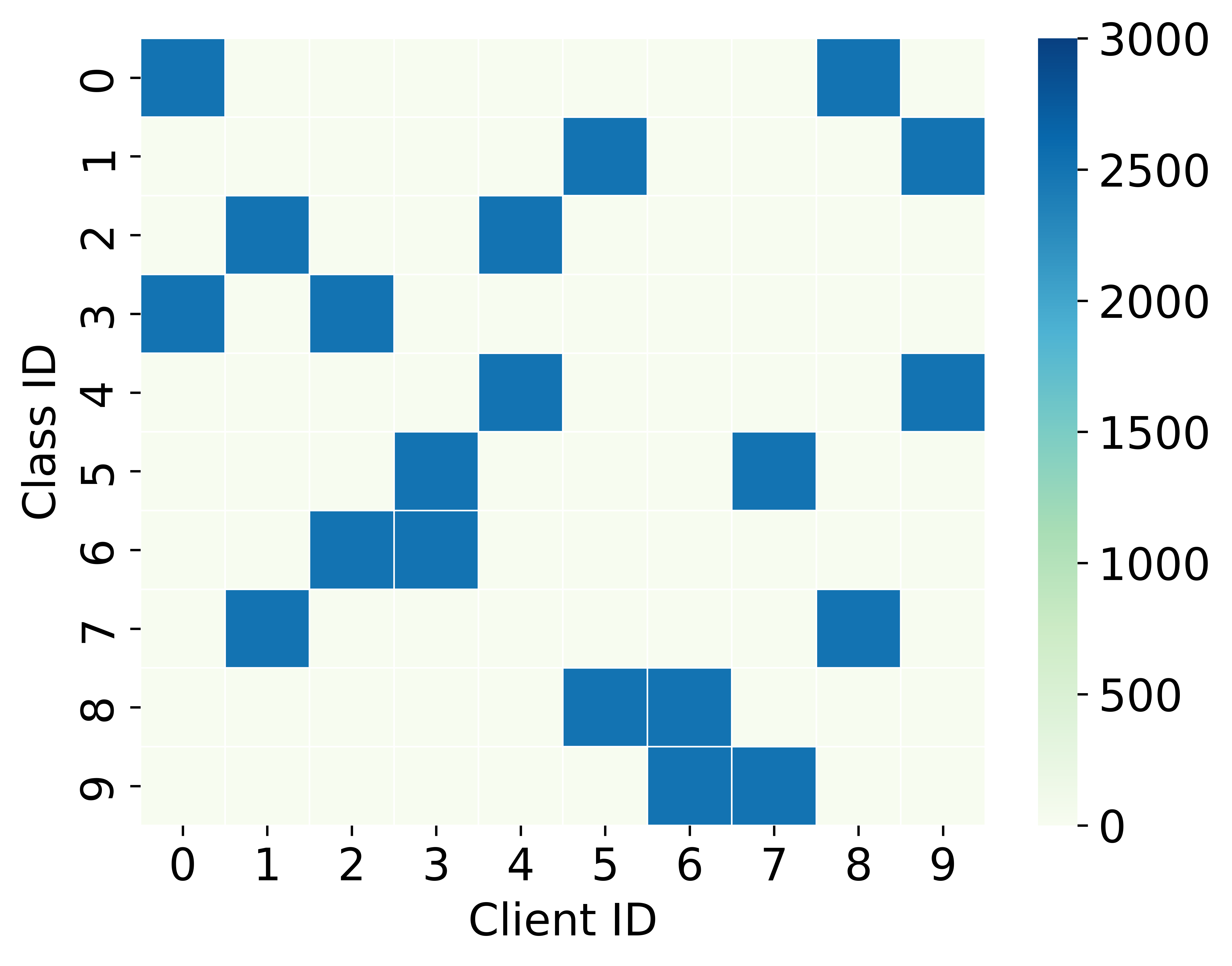}    \label{b}}
        \hfill
        \subfloat[]{\centering\includegraphics[width=2.3in]{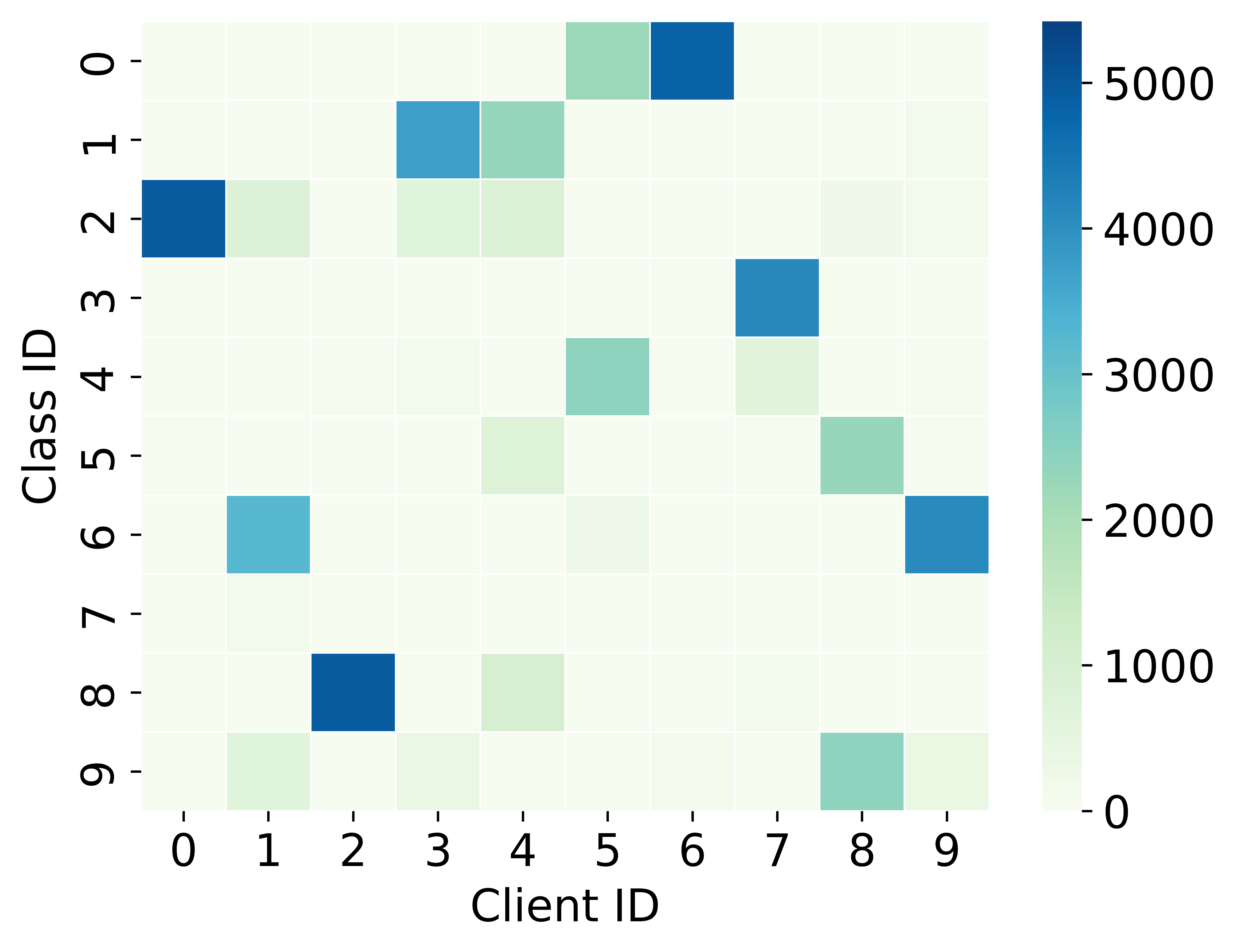}   \label{c}}
    \end{minipage}
    \caption{(a) Distribution of data under the iid setting. (b) Data distribution under the non-iid setting ($\xi$ = 2). 
             (c) Data distribution under the non-iid setting ($\alpha$ = 0.1). The color bar indicates the number of data 
             samples. Each rectangle represents the number of data samples for a specific category in the device.}
    \label{fig:distribution}
\end{figure*}

    In this experiment, we utilize two types of neural networks: the multi-layer perceptron (MLP) and the convolutional neural network (CNN). The MLP 
architecture comprises two hidden layers, each followed by a ReLU activation function, and a classification layer. The MLP model contains a total of 
199210 trainable parameters. This model is employed for training MNIST classification tasks. The CNN network consists of 3 convolutional layers and 
two fully connected layers, with 128420 parameters in total. Both iid and non-iid approaches are used for data classification. For the iid setting, 
we divide the training set randomly and equally to all devices. For the non-iid setting, we will use both pathological and Dirichlet distributions 
for the data partitioning. The pathological distribution is divided by sorting the labels according to the training set, and slicing these labels 
into equal-sized shares, and then the device randomly selects $\xi$ shares from them. Therefore, in the pathological distribution setting, most of 
the devices contain only $\xi$ categories, and a smaller value of $\xi$ represents a higher degree of data heterogeneity. In the Dirichlet partitioning 
approach, we sample $p_i\sim\operatorname{Dir}_K(\alpha)$ and assign a $p_{i,k}$ proportion of the samples from class $i$ to device $k$. In Figure 
\ref{fig:distribution}, we give examples of the partitioning of the pathological and Dirichlet distributions on the CIFAR-10 dataset. For FedSR and 
HierAVG, We assume that the devices participating in training are uniformly distributed across edge servers, and that a device can select a nearby 
edge server to participate in each round of training. We can notice that when the data division of the devices is pathologically distributed, 
$\sum_{m=1}^M\frac{|D_m|^2}{|D|^2}$ is 1/M since the number of local data of each device is the same. therefore when M is greater than 2, our data 
division is satisfying the convergence condition of FedSR. When the data distribution of the devices is Dirichlet distribution, we have counted 
$\sum_{m=1}^M\frac{|D_m|^2}{|D|^2}$ and also from the figure \ref{fig:weight} we can see that $\sum_{m=1}^M\frac{|D_m|^2}{|D|^2}$ is also satisfying 
the convergence condition.

\begin{figure}[!h]
    \centering
    \includegraphics[width=0.4\textwidth]{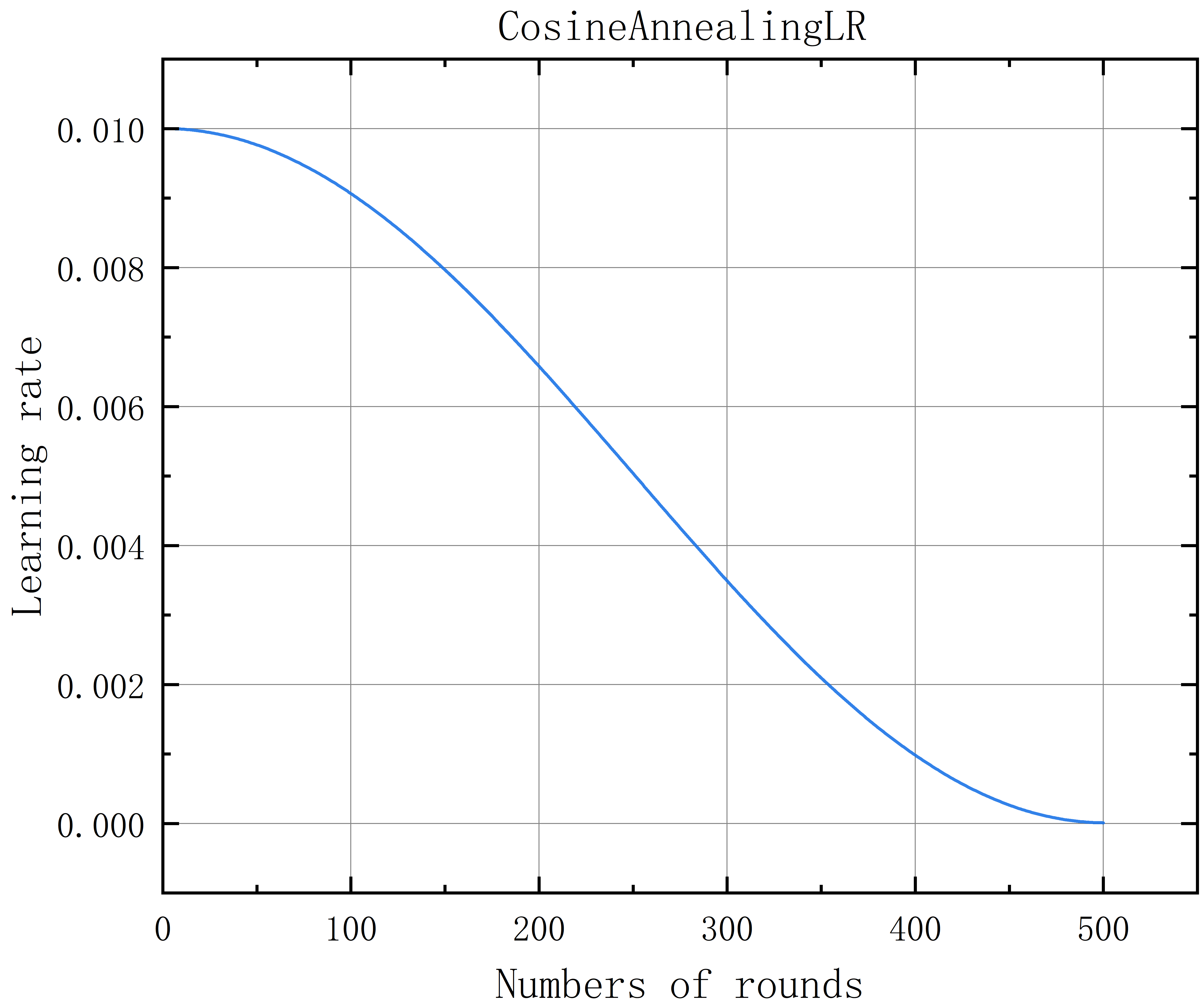}
    \caption{cosine lr decay}
    \label{fig:lr_decay}
\end{figure}

\subsection{Accuracy Comparison}
\begin{table}[h]
    \captionsetup{justification=justified}
    \caption{The performance of different methods on different tasks and different data distribution.}
    \resizebox{\linewidth}{!}{
        \begin{tabular}{@{}ccccccc@{}}
            \toprule 
            Task & Method & IID & $\xi=4$ & $\xi=2$ & $\alpha=0.3$ & $\alpha=0.1$\\
            \midrule
            \multirow{6}{*}{\rotatebox{0}{\shortstack{MNIST}}}
            & Centralized 			        & \textcolor[RGB]{43,213,77}{\textbf{97.53\%}} & \textcolor[RGB]{43,213,77}{\textbf{97.53\%}}
                                            & \textcolor[RGB]{43,213,77}{\textbf{97.53\%}} & \textcolor[RGB]{43,213,77}{\textbf{97.53\%}} 
                                            & \textcolor[RGB]{43,213,77}{\textbf{97.53\%}}\\
            & FedAvg 	                    & 97.64\% & 97.38\% & 93.52\% & 97.50\% & 96.87\%\\
            & FedProx 	                    & 97.60\% & 97.40\% & 93.46\% & 97.43\% & 96.83\%\\
            & MOON 	                        & 97.74\% & 97.30\% & 92.98\% & 97.59\% & 96.62\%\\
            & HierAVG 	                    & 97.69\% & 97.46\% & 95.19\% & 97.53\% & 97.25\%\\
            & ring-optimization             & 97.07\% & 97.08\% & 96.98\% & 97.41\% & 97.17\%\\
            & FedSR 	                    & \textbf{97.71\%}  & \textbf{97.45\%}  & \textbf{97.83\%} & \textbf{97.62\%} & \textbf{97.54\%}\\
            \midrule
            \multirow{6}{*}{\rotatebox{0}{\shortstack{CIFAR-10}}}
            & Centralized 			        & \textcolor[RGB]{43,213,77}{\textbf{81.02\%}} & \textcolor[RGB]{43,213,77}{\textbf{81.02\%}} 
                                            & \textcolor[RGB]{43,213,77}{\textbf{81.02\%}} & \textcolor[RGB]{43,213,77}{\textbf{81.02\%}} 
                                            & \textcolor[RGB]{43,213,77}{\textbf{81.02\%}}\\
            & FedAvg 	                    & 78.20\% & 75.13\% & 68.33\% & 76.99\% & 72.83\%\\
            & FedProx 	                    & 78.24\% & 75.02\% & 68.36\% & 76.91\% & 72.84\%\\
            & MOON 	                        & 79.22\% & 74.80\% & 69.18\% & 76.23\% & 71.41\%\\
            & HierAVG 	                    & 79.47\% & 76.10\% & 69.82\% & 77.78\% & 73.93\%\\
            & ring-optimization             & \textbf{80.90\%}  & \textbf{80.06\%}  & \textbf{79.26\%} & \textbf{80.93\%} & \textbf{78.66\%}\\
            & FedSR 	                    & 80.30\% & 79.59\% & 77.58\% & 79.48\% & 78.10\%\\
            \midrule
            \multirow{6}{*}{\rotatebox{0}{\shortstack{Fashion-MNIST}}}
            & Centralized 			        & \textcolor[RGB]{43,213,77}{\textbf{92.84\%}} & \textcolor[RGB]{43,213,77}{\textbf{92.84\%}}
                                            & \textcolor[RGB]{43,213,77}{\textbf{92.84\%}} & \textcolor[RGB]{43,213,77}{\textbf{92.84\%}}
                                            & \textcolor[RGB]{43,213,77}{\textbf{92.84\%}}\\
            & FedAvg 	                    & 92.62\% & 90.85\% & 85.23\% & 91.49\% & 90.20\%\\
            & FedProx 	                    & 92.54\% & 90.98\% & 85.38\% & 91.71\% & 90.09\%\\
            & MOON 	                        & 92.44\% & 90.67\% & 85.81\% & 90.99\% & 88.90\%\\
            & HierAVG 	                    & 92.48\% & 90.81\% & 86.64\% & 91.99\% & 90.51\%\\
            & ring-optimization             & \textbf{93.02\%}  & \textbf{92.70\%}  & \textbf{92.31\%} & \textbf{92.64\%} & \textbf{92.27\%}\\
            & FedSR 	                    & 92.93\% & 92.57\% & 92.04\% & 92.24\% & 92.24\%\\
            \midrule
            Task & Method & IID & $\xi=8$ & $\xi=4$ & $\alpha=0.3$ & $\alpha=0.1$\\
            \midrule
            \multirow{6}{*}{\rotatebox{0}{\shortstack{CIFAR-100}}}
            & Centralized 			        & \textcolor[RGB]{43,213,77}{\textbf{49.04\%}} & \textcolor[RGB]{43,213,77}{\textbf{49.04\%}}
                                            & \textcolor[RGB]{43,213,77}{\textbf{49.04\%}} & \textcolor[RGB]{43,213,77}{\textbf{49.04\%}} 
                                            & \textcolor[RGB]{43,213,77}{\textbf{49.04\%}}\\
            & FedAvg 	                    & 43.49\% & 34.18\% & 29.99\% & 42.01\% & 39.95\%\\
            & FedProx 	                    & 42.84\% & 33.92\% & 30.54\% & 42.89\% & 39.34\%\\
            & MOON 	                        & 43.41\% & 33.49\% & 29.58\% & 42.18\% & 38.97\%\\
            & HierAVG 	                    & 36.31\% & 27.96\% & 23.84\% & 34.87\% & 32.51\%\\
            & ring-optimization             & \textbf{49.70\%}  & \textbf{46.14\%}  & 41.64\% & \textbf{49.53\%} & \textbf{48.56\%}\\
            & FedSR 	                    & 47.94\% & 45.53\% & \textbf{44.44\%}  & 48.94\% & 47.07\%\\
            \bottomrule
        \end{tabular}
    }
    \label{tab:IID_and_Non-IID}
\end{table}

\label{sec:Accuracy Comparison}
    In our first experiment, we compare the performance of FedSR with four SOTA federated learning algorithms across various datasets and in 
the case of iid setting and non-iid setting ($\xi=4$, $\xi=2$, $\alpha=0.3$ and $\alpha=0.1$). We evaluate the performance of the global model on 
four image classification tasks: MNIST, FashionMNIST, CIFAR-10, and CIFAR-100. The MLP will be trained on the MNIST dataset, and the CNN will be 
trained on the FashionMNIST, CIFAR-10, and CIFAR-100 datasets. We adopt the SGD optimizer to optimize the model. the momentum of SGD is set to 0.5 
and the batchsize is set to 32. The initial learning rate of the SGD algorithm is set to 0.01 and gradually decays in a cosine function pattern as 
the number of training rounds increases. Eventually, the learning rate decays to $1 \times 10^{-5}$ in the final training round. For example, if the 
training round is 500 rounds, the learning rate decay of the device is shown in Fig. \ref{fig:lr_decay}. The number of devices is set to K=20 and all 
devices participate in the training in each round. For FedAvg, FedProx, and MOON, the number of local update rounds is set to E=5 for all tasks. For 
FedSR and HierFAVG, we divided the devices participating in the training into 5 clusters equally, with the local epoch of each device set to 1. In 
each round, the server aggregates models from all clusters after each cluster has completed 5 iterations. We ensures the computational resources spend 
by devices in all methods are equal in each round.
\par
    The test accuracy of all approaches is presented in Table \ref{tab:IID_and_Non-IID}. We can find that in most cases FedSR achieves higher model 
accuracy in both iid and non-iid settings compared to FedAvg, FedProx, MOON, and HierAVG, and is slightly lower than ring-optimization approach. In 
non-iid seting ($\xi=4$), we can note that the model performance of FedSR and ring-optimization on MNIST and FashionMNIST is still close to that of 
the iid case, however the model performance of FedAvg, FedProx, MOON, and HierAVG degrades dramatically. And on CIFAR-10 and CIFAR-100, FedSR and ring-
optimization also has remarkable accuracy in iid and non-iid cases. In Figure. \ref{fig:IID_and_Non_IID}, we show the global model test accuracy curves 
for each algorithm. It is worth noting that the model accuracy of FedSR is slightly lower than that of ring-optimization, however FedSR has faster convergence 
rates and ring-optimization takes more training time due to the serial training manner.
\subsection{Number of Local Epochs and Ring Epochs}
\begin{figure}[!h]
    \centering
    \begin{minipage}[b]{\linewidth}
        \centering
        \subfloat[]{\centering\includegraphics[width=3in]{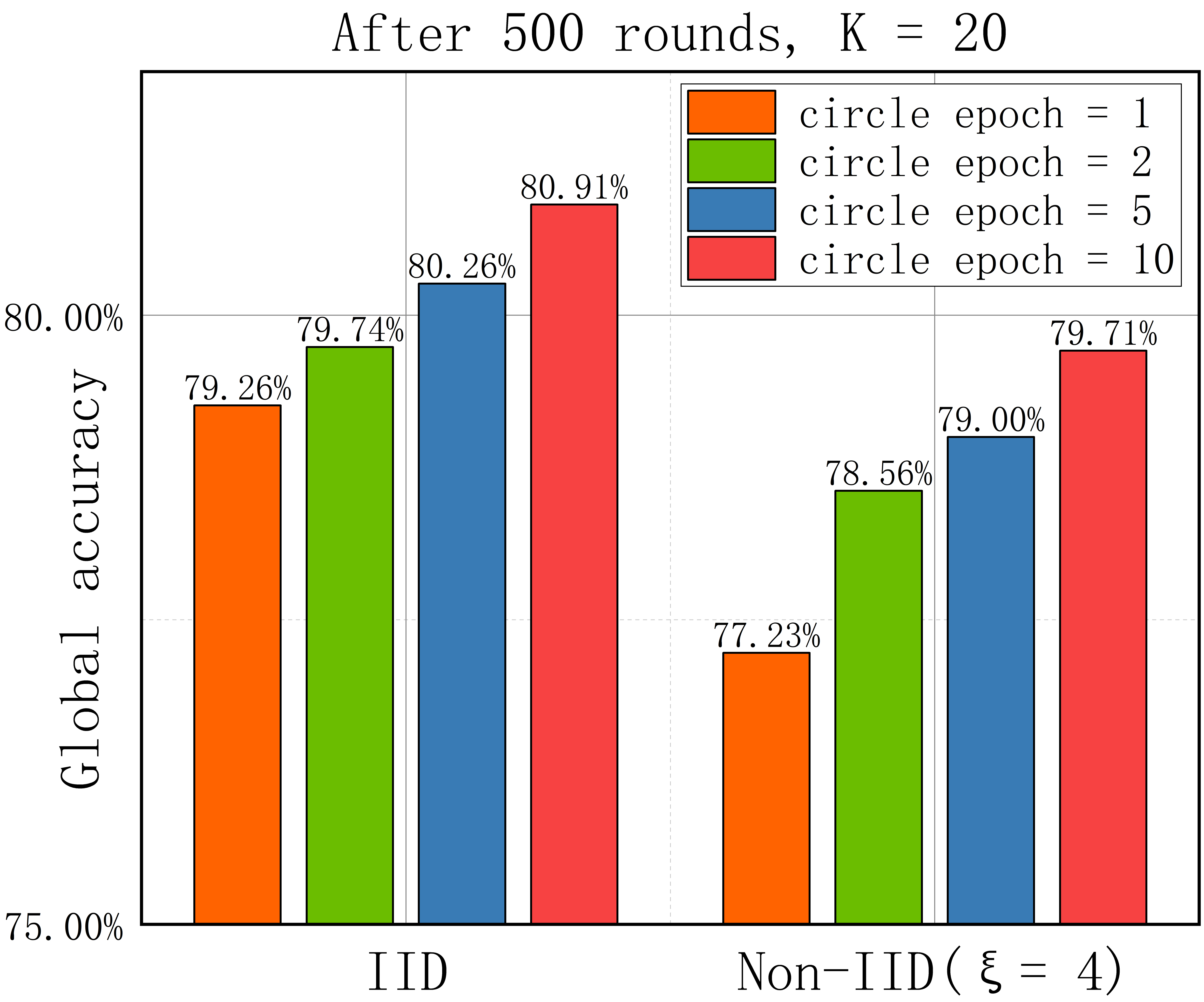}   \label{a}}
        \hfill
        \subfloat[]{\centering\includegraphics[width=3in]{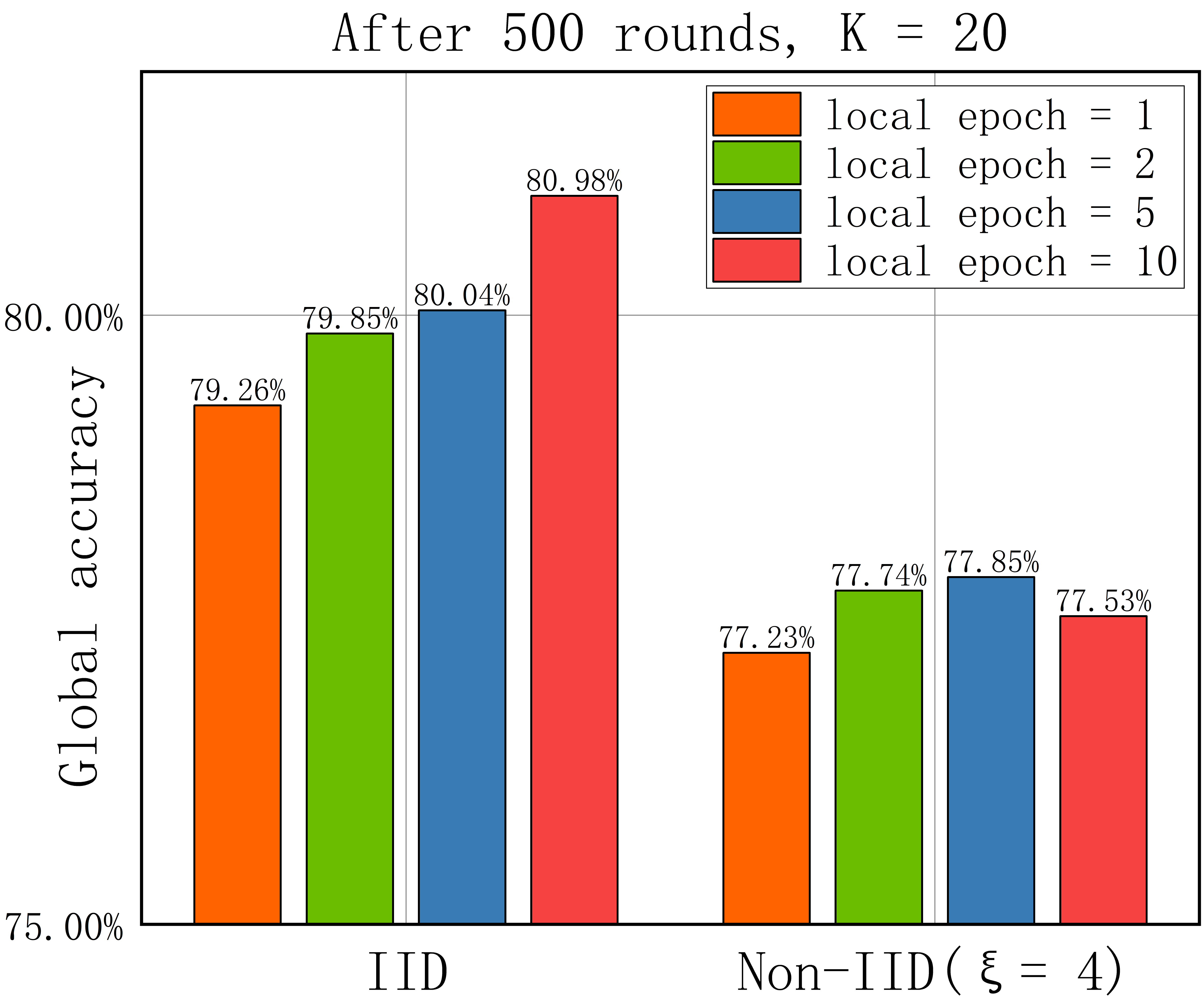}    \label{b}}
    \end{minipage}
    \caption{The impact of local epoch and circle epoch on FedSR}
    \label{fig:circle_and_local_epoch}
\end{figure}

    To analyze the impact of the number of local updates and ring optimization rounds on FedSR, we conducted experiments on the CIFAR-10 dataset. 
For the impact of local updates, we set the number of ring optimization epoch to 1 and varied the number of local epoch to 1, 2, and 5. For the 
impact of ring optimization epoch, we set the number of local updates to 1 and varied the number of ring optimization epoch to 1, 2, and 5. We 
use the SGD optimizer to optimize the model, and train for a total of 500 rounds. The momentum of SGD is set to 0.5, and the batch size is set 
to 32. The initial learning rate of the SGD algorithm is set to 0.01, and it decays in a cosine manner. The number of devices was set to 20, and 
all devices participated in training in each round. In FedSR, we divide the 5 clusters into groups of 4 devices each. The device dataset partitioning 
followed the pathological distribution with $\xi=4$. The experimental results are depicted in Figure \ref{fig:circle_and_local_epoch}.
    In iid setting, we can see from Figure \ref{fig:circle_and_local_epoch} that increasing the number of epochs for local training and the 
number of epochs for ring optimization both increase the training accuracy of the model for the same computation burden. However, in non-iid 
setting, increasing the number of epochs of loop optimization results in better model performance than increasing the number of epochs of local 
updates. Increasing the number of rounds of ring optimization allows for more frequent information exchange among devices, thus preventing the 
model from biasing towards local optima. In non-iid setting ($\xi=4$), we can observe that when devices perform 10 local update epochs, the model's 
performance is worse compared to when they perform 1, 2, or 5 local update epochs. This is because increasing local update rounds can lead to 
the model being biased towards local optima, thereby impacting the generalization ability of the global model.
\subsection{The impact of the number of ring clusters}
\begin{figure}[!h]
    \centering
    \includegraphics[width=0.4\textwidth]{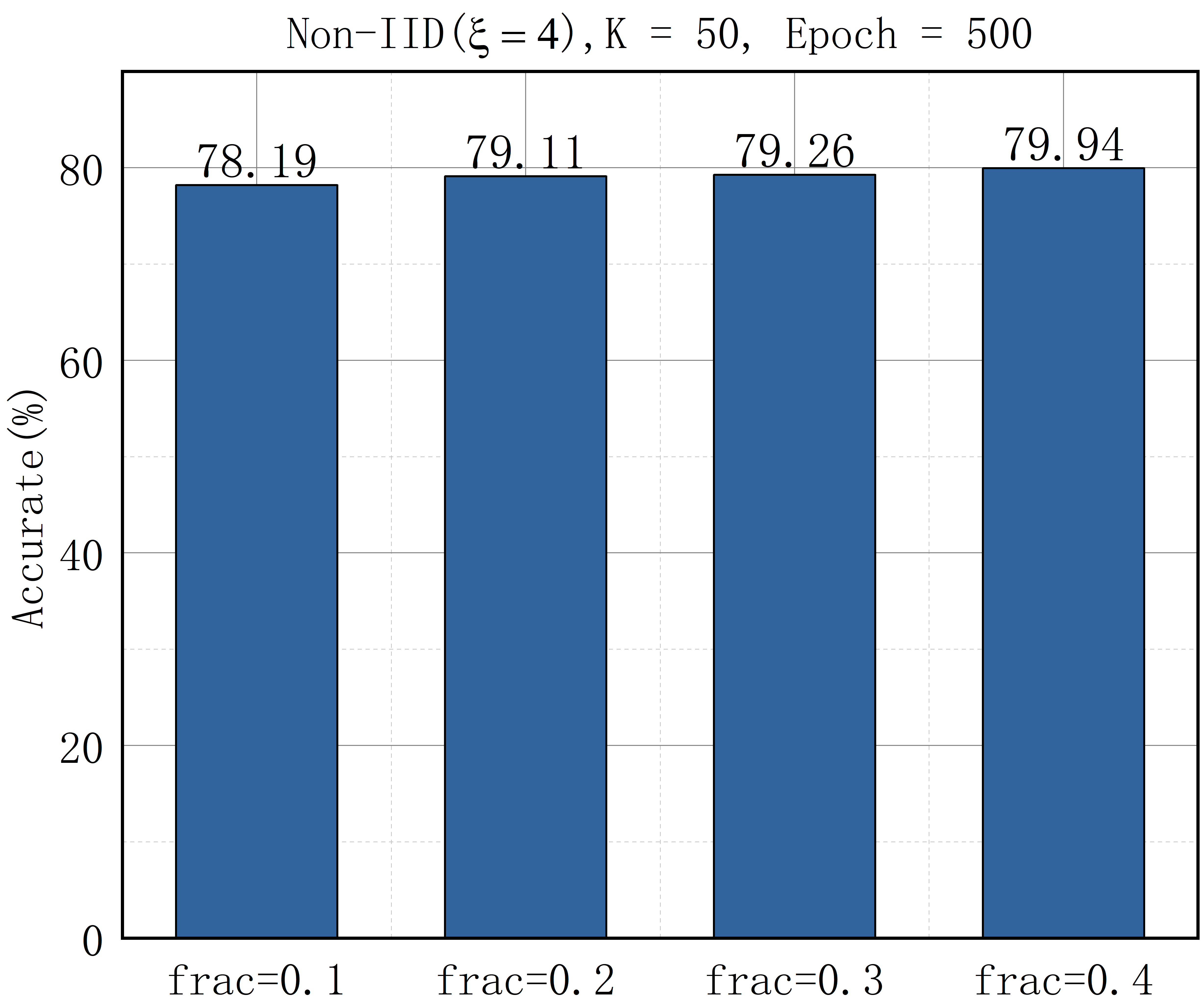}
    \caption{The impact of the number of ring clusters for FedSR}
    \label{fig:frac}
\end{figure}
    We also conducted experiments on the impact of the number of devices in different device cluster on iid and non-iid ($\xi=4$) setting. On 
the CIFAR-10 dataset, we set the number of participating devices in federated learning to 50. Each device performed 1 local training round, and 
the number of ring optimization was set to 5. We varied the proportion of devices in the ring cluster as a percentage of the total number of devices, 
with values of 0.1, 0.2, 0.3, 0.4. The experimental results are depicted in Figure \ref{fig:frac}. As the number of devices in the ring cluster 
increases, we can observe an improvement in the performance of the global model. This is because the devices in each ring topology have access to 
the knowledge of more devices. Although increasing the number of devices in the ring cluster can improve the performance of the model, it also brings 
more training time. So FedSR need to make a trade-off between performance and training time.
\subsection{Communication Cost}
    We calculated the communication cost of all algorithms to achieve the same model accuracy on MNIST, FashionMNIST, CIFAR-10, and CIFAR-100 datasets 
with the same experimental settings as in section \ref{sec:Accuracy Comparison}. The experiment result is shown in the Table. \ref{tab:communication_cost}. 
We can find that FedSR is not superior in time overhead compared to model-aggregated federated learning algorithms, however, it can achieve better model 
performance with less communication overhead.
\begin{table}[!h]
    \captionsetup{justification=justified}
    \caption{The cost of communication to achieve the target accuracy in the setting K = 20 and the M represents the traffic for transferring one model.}
    \resizebox{\linewidth}{!}{
        \begin{tabular}{@{}ccccccc@{}}
            \toprule 
                                    & MNIST($\xi=2$)  & FashionMNIST($\xi=2$)  & CIFAR-10($\xi=2$)   & CIFAR-100($\xi=4$)\\
            \midrule
            Target accuracy         &  90\%         & 80\%          & 67\%          & 33\%\\
            \midrule
            FedAvg                  & 4680M          & 3200M              & 9760M              & 14720M          \\
            FedProx                 & 4600M          & 2800M              & 9840M              & 14960M          \\
            MOON                    & 4040M          & 3000M              & 9800M              & —               \\
            HierAVG                 & 3410M          & 3410M              & 14520M             & 29150M          \\
            ring-optimization       & \textbf{260M}  & 3120M              & 7080M              & 9140M           \\
            FedSR                   & 660M           & \textbf{1980M}     & \textbf{6160M}     & \textbf{5390M}  \\
            \bottomrule
        \end{tabular}
    }
    \label{tab:communication_cost}
\end{table}
\subsection{Scalability}
\begin{table}[!h]
    \captionsetup{justification=justified}
    \caption{The accuracy with 100 devices in different participation rate on CIFAR-10 in non-iid setting ($\xi=2$).}
    \resizebox{\linewidth}{!}{
        \begin{tabular}{@{}ccccccc@{}}
            \toprule
            \multirow{2}{*}{Method} &\multicolumn{2}{c}{sample fraction=0.2} &\multicolumn{2}{c}{sample fraction=0.4}\\
            & epochs=250 & epochs=500 & epochs=250 & epochs=500\\
            \midrule
            FedAvg                & 62.08\% & 68.31\% & 62.62\% & 68.85\%\\
            FedProx               & 62.74\% & 68.48\% & 64.65\% & 68.92\%\\
            MOON                  & 64.52\% & 69.26\% & 66.17\% & 70.12\%\\
            HierAVG               & 65.14\% & 69.18\% & 62.08\% & 69.47\%\\
            ring-optimization     & 69.19\% & 76.48\% & \textbf{72.66\%} & \textbf{78.46\%}\\
            FedSR                 & \textbf{70.05\%} & \textbf{77.30\%}  & 71.56\% & 76.73\%\\
            \bottomrule
        \end{tabular}
    }
    \label{tab:scalability}
\end{table}
    To show the scalability of FedSR, we experimented with a larger number of devices on the CIFAR-10. We set the number of devices to 100, the 
participation rate of devices per round to 0.2 and 0.4, and the number of devices per ring cluster to 4 for FedSR, respectively. The other settings 
are the same as in section \ref{sec:Accuracy Comparison}. The experiment results are shown in Table \ref{tab:scalability}. From the Table \ref{tab:scalability}, 
we can see that FedSR has a superior performance compared to the other centralized federated learning approaches in the case of large device participation.
Although ring-optimization can achieve higher accuracy than FedSR, it takes a lot of time, due to its serial training approach, and is not practical.

    \section{Conclusion}
    In this paper, we consider federated learning in the Internet of Things (IoT) and propose a 
semi-decentralized hierarchical federated learning framework. In order to reduce the impact of 
data heterogeneity, we introduce the incremental subgradient optimization algorithm in the training 
of ring cluster. And the semi-decentralized hierarchical federated learning framework can effectively 
reduce the communication pressure on the cloud server by dividing the participating clients into 
different ring clusters for training. Meanwhile, We conducted experiments on MNIST, FashionMNIST, 
CIFAR-10, and CIFAR-100 datasets. The results of our experiments show that the method proposed in 
this paper has higher model accuracy in non-iid setting and lower communication cost compared to 
previous federated learning algorithms. In the future, we plan to take the system heterogeneity 
problem into account to further optimize training efficiency.

    \bibliography{ref.bib}
    \bibliographystyle{ieeetr}
\end{document}